\documentclass{cta-author}%%%%where cta-author is the template name

{}
{}
{}
\usepackage[table]{xcolor}% http://ctan.org/pkg/xcolor
\usepackage{color}
\usepackage{graphicx}
\usepackage{times}
\usepackage[colorlinks=true, allcolors=blue]{hyperref}
\usepackage{amsmath}
\usepackage{pifont}
\usepackage{mathtools}
\usepackage{rotating}
\usepackage{float}

\begin{document}

\title{Learning to Recognize 3D Human Action from A New Skeleton-based Representation Using Deep Convolutional Neural Networks}

\author{\au{Huy-Hieu~Pham $^{1,2,*}$, }%%% First author
\au{Louahdi~Khoudour $^1$, }%%% Second author
\au{Alain~Crouzil $^2$, }%%% Third author
\au{Pablo~Zegers $^3$, }%%%  Fourth author
\au{Sergio~A. Velastin $^{4,5,6}$}%%% Fifth author
}

\address{\add{1}{Cerema, Equipe-projet STI, 1 Avenue du Colonel Roche, 31400, Toulouse, France}
\add{2}{Institut de Recherche en Informatique de Toulouse (IRIT), Universit\'e de Toulouse, UPS, 31062 Toulouse, France}
\add{3}{Aparnix, La Gioconda 4355, 10B, Las Condes, Santiago, Chile}
\add{4}{Department of Computer Science, Applied Artificial Intelligence Research Group, University Carlos III de Madrid, 28270 Madrid, Spain}
\add{5}{School of Electronic Engineering and Computer Science, Queen Mary University of London, UK}
\add{6}{Cortexica Vision Systems Ltd., London, UK}
\email{{\color{blue} \underline{huy-hieu.pham@cerema.fr}}}}

\begin{abstract}
Recognizing human actions in untrimmed videos is an important challenging task. An effective 3D motion representation and a powerful learning model are two key factors influencing recognition performance. In this paper we introduce a new skeleton-based representation for 3D action recognition in videos. The key idea of the proposed representation is to transform 3D joint coordinates of the human body carried in skeleton sequences into RGB images via a color encoding process. By normalizing the 3D joint coordinates and dividing each skeleton frame into five parts, where the joints are concatenated according to the order of their physical connections, the color-coded representation is able to represent spatio-temporal evolutions of complex 3D motions, independently of the length of each sequence. We then design and train different Deep Convolutional Neural Networks (D-CNNs) based on the Residual Network architecture (ResNet) on the obtained image-based representations to learn 3D motion features and classify them into classes. Our method is evaluated on two widely used action recognition benchmarks: MSR Action3D and NTU-RGB+D, a very large-scale dataset for 3D human action recognition. The experimental results demonstrate that the proposed method outperforms previous state-of-the-art approaches whilst requiring less computation for training and prediction.
\end{abstract}

\maketitle

\section{Introduction}
\label{sec:Introduction}
Human Action Recognition (HAR) \cite{Ranasinghe2016ARO} is an important topic in machine vision. HAR methods have been widely applied in building electronic systems based on machine intelligence such as intelligent surveillance \cite{Niu2004HumanAD,Lin2008HumanAR}, human-machine interaction \cite{Pickering2007ARS,Sonwalkar2015HandGR}, health
care \cite{pansiot2007ambient}, and so on. Although significant progress has been made in the last few years, HAR is still a challenging task due to many obstacles, \textit{e.g.} camera viewpoint, occlusions, or large intra-class variations \cite{Poppe2010ASO}.\\[0.2cm]
\hspace*{0.3cm} There are two main types of intelligence that
are artificial intelligence and machine intelligence. HAR can be considered as an electronic system based on machine intelligence. The first stage of any HAR system is data acquisition. Nowadays, a variety of electronic imaging systems can be used for this task such as traditional optical cameras, RGB-D sensors, Thermal Infrared (IR) sensors, and Synthetic Aperture Radar (SAR) sensors. For instance, IR sensors have been exploited in HAR \cite{akula2018deep}. These sensors are able to generate images based on the heat radiated by the human body and to work independently of lighting conditions. SAR imaging has also been widely used for detecting and analyzing human activities \cite{gurbuz2007detection}. This type of sensor is able to operate far away from potential targets and functions during the daytime as well as nighttime, under all weather conditions. In particular, SAR Image Segmentation (SIS) techniques \cite{akbarizadeh2012new,tirandaz2016unsupervised,farbod2018optimized} can segment humans from other components (\textit{e.g.} objects, background) in the scene. As a result, the use of SAR sensors offers big advantages for human action recognition, in particular for human target detection and identification in a complex environment such as in military applications. \\[0.2cm]
\hspace*{0.3cm} Traditional studies on HAR mainly focus on the use of RGB sequences provided by 2D cameras. These approaches typically recognize actions based on hand-crafted local features such as Cuboids \cite{Dollr2005BehaviorRV}, HOG/HOF \cite{Laptev2008LearningRH}, HOG-3D \cite{Klser2008ASD}, which are extracted from the appearance and movements of human body parts. However, it is very hard to fully capture and model the spatial-temporal information of human action due to the absence of 3D structure from the scene. Therefore, RGB-D cameras are becoming one of the most commonly-used sensors for HAR \cite{zhang2016rgb,ye2013survey,ijjina2017human}. Recently, the rapid development of cost-effective and easy-to-use depth cameras, \textit{e.g.} Microsoft Kinect $^{\text{TM}}$sensor \cite{Cruz2012KinectAR,Han2013EnhancedCV}, ASUS Xtion PRO \cite{ASUS}, or Intel$^{\tiny{\textregistered}}$ RealSense$^{\text{TM}}$ \cite{Intel} has helped dealing with problems related to 3D action recognition. In general, depth sensors are able to provide detailed information about the 3D structure of human body. Therefore, many approaches have been proposed for 3D HAR based on data provided by depth sensors such as RGB sequences, depth, or combining these two data types (RGB-D). Furthermore, these devices have integrated real-time skeleton tracking algorithms \cite{Shotton2011RealtimeHP} that provide high-level information about human motions in a 3D space. Thus, exploiting skeletal data for 3D HAR is an active research topic in computer vision \cite{Wang2012MiningAE,Xia2012ViewIH,Chaudhry2013BioinspiredD3,Vemulapalli2014HumanAR,Ding2016ProfileHF}.\\[0.2cm]
\hspace*{0.3cm} In recent years, approaches based on Convolutional Neural Networks (CNNs) have achieved outstanding results in many image recognition tasks \cite{Krizhevsky2012ImageNetCW,Karpathy2014LargeScaleVC,Szegedy2016RethinkingTI}. After the success of AlexNet \cite{Krizhevsky2012ImageNetCW}, a new direction of research has been opened for designing and optimizing higher performing CNN architectures. As a result, there are substantial evidences that seem to show that the learning performance of CNNs may be significantly improved by increasing the number of hidden layers \cite{Simonyan2014VeryDC,Szegedy2015GoingDW,Telgarsky2016BenefitsOD}. In the literature of HAR, many studies have indicated that CNNs have the ability to learn complex motion features better than hand-crafted approaches. However, most previous works have just focused on exploring simple CNNs such as AlexNet \cite{Krizhevsky2012ImageNetCW} and have not exploited the potential of recent state-of-the-art very deep CNN architectures (D-CNNs), \textit{e.g.} Residual Networks (ResNet) \cite{He2016DeepRL}. In addition, most existing CNN-based approaches limit themselves to using RGB-D sequences as the input to learning models. Although RGB-D images are informative for understanding human action, the computation complexity of learning models increases rapidly when the dimension of the input features is large. Therefore, representation learning models based on RGB-D modality become more complex, slower and less practical for solving large-scale problems as well as real-time applications.\\[0.2cm]
\hspace*{0.3cm} Different from previous works, to take full advantages of 3D skeletal data and the learning capacity of D-CNNs, this paper proposes an end-to-end deep learning framework for 3D HAR from skeleton sequences. We focus on solving two main issues: First, using a simple skeleton-to-image encoding method to transform the 3D coordinates of the skeletal joints into RGB images. The encoding method needs to ensure that the image-based representation of skeleton sequences is able to effectively represent the spatial structure and temporal dynamics of the human action. Second, we design and train different D-CNNs based on Residual Networks (ResNets) \cite{He2016DeepRL} -- a recent state-of-the-art CNN for image recognition, to learn and classify actions from the obtained image-coded representations. \\
\hspace*{0.3cm} This paper is an extended version of our work published in the 8th International Conference of Pattern Recognition Systems (ICPRS 2017) \cite{iet:/content/conferences/10.1049/cp.2017.0154}. It is part of our research project about investigating and developing a low-cost system based on RGB-D data provided by depth sensors for understanding human actions and analyzing their behaviors in indoor environments such as inside home, offices, residential buildings, buses, trains, etc. In general, three observations motivate our exploration of using ResNet \cite{He2016DeepRL} for 3D HAR from skeleton sequences, including: \textbf{(1)} human actions can be correctly
represented as the movements of skeletons \cite{Vemulapalli2014HumanAR} and the spatial-temporal dynamics of skeletons can be transformed into 2D image structure, which can be effectively learned by D-CNNs; \textbf{(2)} skeletal data is high-level information with much less complexity than RGB-D sequences, this advantage makes our action learning model much simpler and faster; \textbf{(3)} many evidences \cite{Szegedy2015GoingDW,Telgarsky2016BenefitsOD} show that deeper convolutional model, especially ResNet \cite{He2016DeepRL} can boost the learning accuracy in image recognition tasks.\\[0.2cm]
\hspace*{0.3cm} We evaluate the proposed method on two benchmark skeleton datasets (\textit{i.e.} MSR Action3D \cite{Li2010ActionRB} and NTU-RGB+D \cite{Shahroudy2016NTURA}). Experimental results confirmed the above statements since our method achieveds state-of-the-art performance compared with the existing results using the same evaluation protocols. Furthermore, we also indicate the effectiveness of our learning framework in terms of computational complexity.\\[0.2cm]
\hspace*{0.3cm} In summary, two main contributions of this paper include:\\
\\
$\bullet$ First, we introduce a new skeleton-based representation (Sec. \ref{image-based-representation}) and an end-to-end learning framework\footnote{Codes and pre-trained models will be shared to the community at \url{https://github.com/huyhieupham/} after publication.} (Sec. \ref{residual-networks})  based on D-CNNs to learn the spatio-temporal evolutions of 3D motions and then to recognize human actions;\\
\\
$\bullet$ Second, we show the effectiveness of our method on HAR tasks by achieving state-of-the-art performance on two benchmark datasets, including the most challenging skeleton benchmark currently available for 3D HAR \cite{Shahroudy2016NTURA} (Sec. \ref{sect:5}).\\
\\
The advantage of the proposed method is that it has high computational efficiency (Sec. \ref{prediction-time}). Moreover, this approach is general and can be easily applied to other time-series problem, \textit{e.g.} recognizing human actions with mobile devices with integrated inertial sensors. \\[0.2cm]
\hspace*{0.3cm} The rest of the paper is organized as follows: Section~\ref{sect:2} discusses related works. In Section~\ref{sect:3}, we present the details of our proposed method, including the color-encoding process from skeletons to RGB images and deep learning networks. Datasets and experiments are described in Section~\ref{sect:4}. Experimental results are reported in Section~\ref{sect:5} with a detailed analysis of computational efficiency. Finally, Section~\ref{sect:7} concludes the paper with a discussion on the future work.
\section{Related Work}
\label{sect:2}
Skeleton-based Action Recognition (SBAR) using depth sensors has been widely studied in recent years. In this section, we briefly review existing SBAR methods, including two main categories that are directly related to our work. The first category is approaches based on hand-crafted local features of skeletons. The second is approaches based on deep learning networks, especially Recurrent Neural Networks with Long Short Term Memory (RNN-LSTMs).\\
\\
\textit{\textbf{Approaches based on hand-crafted local features}}: Human motion can be considered as a spatio-temporal pattern. Many researchers have built hand-crafted feature representations for SBAR and then used temporal models for modeling 3D human motion. For instance, Wang \textit{et al}. \cite{Wang2012MiningAE} represented the human motion from skeletal data by means of the pairwise relative positions of the key joints. The authors then employed Hidden Markov Model (HMM) \cite{Yoon2009HiddenMM} for modeling temporal dynamics of actions. A similar approach has been presented by Lv \textit{et al}. \cite{Lv2006}  where the 3D joint position trajectories have been mapped into feature spaces. The dynamics of actions was then learned by one continuous HMM. Xia \textit{et al}. \cite{Xia2012ViewIH} proposed the use of histogram-based representation of human pose, and then actions were classified into classes by a discrete HMM. Wu and Shao \cite{Wu2014LeveragingHP} also exploited HMM for recognizing actions from high-level features of skeletons. Vemulapalli \textit{et al}. \cite{Vemulapalli2014HumanAR} represented the 3D geometric interactions of the body parts in a Lie Group. These elements were processed with a Fourier Temporal Pyramid (FTP) transformation for modeling temporal evolutions of the original motions. Different from the studies above, Luo \textit{et al}. \cite{Luo2013GroupSA} proposed a new dictionary learning model to learn the spatio-temporal information from skeleton sequences. Although promising results have been achieved from approaches based on hand-crafted local features and probabilistic graphical models, they have some limitations that are very difficult to overcome. \textit{E.g.}, most of these
approaches are data-dependent and require a lot of hand-designing features. HMM-based methods require preprocessing input data in which the skeleton sequences need to be segmented or aligned. Meanwhile, FTP-based approaches can only utilize limited contextual information of actions.\\
\\
\textit{\textbf{Approaches based on deep learning}}: Recurrent Neural Networks with Long Short-Term Memory Network (RNN-LSTMs) \cite{hochreiter1997long,Graves2008SupervisedSL} are able to model the contextual information of the temporal sequences as skeleton data. Thus, many authors have explored RNN-LSTMs for SAR. For instance, Du \textit{et al}. \cite{7298714} proposed an end-to-end hierarchical RNN-LSTM for modeling local motions of a body part in which all skeleton frames were divided into five parts according to the human physical structure. Each part of a skeleton  was then fed into an independent  RNN and then fused to be the inputs of higher layers. The final representation was used for the classification task. Zhu \textit{et al}. \cite{zhu2016co} proposed a LSTM network with a mixed-norm regularization term to cost functions in order to learn the co-occurrence of discriminative features from skeletal data. Liu \textit{et al}. \cite{Liu2016SpatioTemporalLW} introduced a new gating mechanism with LSTM network to analyze the reliability of input skeleton sequences. In another study, Shahroudy \textit{et al}. \cite{Shahroudy2016NTURA} used a part-aware LSTM network in which the memory-cell was divided into sub-cells such that each sub-cell can model long-term contextual representation of a body part. Finally, all these sub-cells were concatenated to the final output. Although RNN-LSTMs are able to model the long-term temporal of motion and experimental results provided that RNN-LSTMs have outperformed many other approaches, RNN-based approaches just consider skeleton sequences as a kind of low-level feature by feeding directly the raw 3D joint coordinates carried in skeletons into the network input. The huge number of input parameters may make RNNs become very complex and easily lead to overfitting due to insufficient training data. Moreover, many RNN-LSTMs act as a classifier and cannot extract high-level features \cite{Sainath2015ConvolutionalLS} for recognition tasks.\\[0.2cm]
\hspace*{0.3cm} Several authors have exploited the feature learning ability of CNNs on skeletal data \cite{wang2016action,hou2016skeleton}. However, such studies mainly focus on the use of complex encoding methods for finding good skeletal representations and learning geometric features carried in skeleton sequences with simple CNN architectures. In contrast, in this paper we concentrate on proposing a new and simple skeleton-based representation and  exploiting the power of D-CNNs for action recognition. Our experiments on two public datasets, including the MSR Action3D dataset \cite{Li2010ActionRB} and the NTU-RGB+D dataset \cite{Shahroudy2016NTURA} show state-of-the-art performance on both datasets. Moreover, in terms of computational cost, our measurement and analyses show that the proposed model is fast enough for many real-time applications.

\section{Proposed Method}
\label{sect:3}
This section presents the proposed method. We first describe the color-encoding process that allows us to represent the spatial structure and temporal dynamics of skeleton sequences as the static structure of a 2D color image. We then review the main idea of ResNet \cite{He2016DeepRL} and propose five different ResNet configurations, including 20-layer, 32-layer, 44-layer, 56-layer, and 110-layer networks to learn motion features carried in the image-based representations and perform action classification.
\subsection{Building image-based representation from skeleton sequences} \label{image-based-representation}
\label{image-based representation}
\begin{figure}[h]
\centering\includegraphics[width=6.5cm,height=5cm]{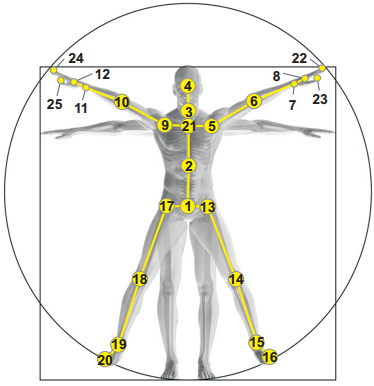}
\caption{The position of 25 joints in the human body extracted by Kinect v2 sensor \cite{Shahroudy2016NTURA}. Skeleton sequences can be recorded frame-by-frame at 30 FPS.}
\label{fig:kinect} 
\end{figure}

\begin{figure}[H]
\centering
\includegraphics[width=9cm,height=2.5cm]{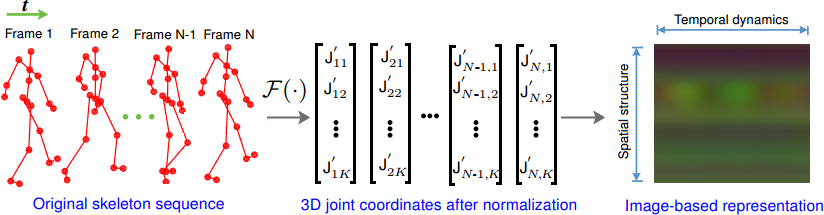} 
\caption{\label{fig:skeleton-rgb} Illustration of the data transformation process. Each skeleton sequence is encoded into a single color image that represents the spatio-temporal evolutions of the motion. Here, $N$ denotes the number of frames in each sequence while $K$ denotes the number of joints in each frame. The value of $K$ depends on each RGB+D dataset. \textit{E.g.}, $K$  is equal to 20 for MSR Action3D dataset \cite{Li2010ActionRB}. Meanwhile, $K$  is equal to 25 for NTU-RGB+D dataset \cite{Shahroudy2016NTURA}. }
\end{figure}

\begin{figure}[h]
\centering
\includegraphics[width=8.5cm,height=3.5cm]{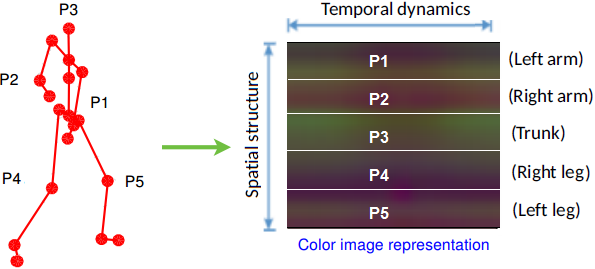} 
\caption{\label{fig:rearrange} Rearranging the order of joints according to the human body physical structure.}
\end{figure}

\begin{figure}[h]
\centering
\includegraphics[width=2.5cm,height=2cm]{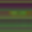} \hspace{0.2cm}
\includegraphics[width=2.5cm,height=2cm]{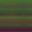} \hspace{0.2cm}
\includegraphics[width=2.5cm,height=2cm]{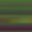} \\
{\footnotesize (a) DrawX \hspace{1.1cm} (b) Forward kick \hspace{0.8cm} (c) Hand catch}\\
\includegraphics[width=2.5cm,height=2cm]{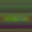} \hspace{0.2cm}
\includegraphics[width=2.5cm,height=2cm]{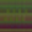} \hspace{0.2cm}
\includegraphics[width=2.5cm,height=2cm]{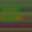} \\
{\footnotesize (d) High throw \hspace{1.2cm} (e) Jogging \hspace{1.1cm} (f) Two hand wave}

\caption{\label{fig:image-representation} The image-based representations obtained from several samples of MSR Action3D dataset \cite{Li2010ActionRB}. In our experiments, all images are resized to $\textbf{32} \times \textbf{32}$ pixels. Best viewed in color.}
\end{figure}

Currently, skeletal data that contains 3D joint coordinates can be obtained from depth cameras via real-time skeleton estimation algorithms \cite{Davison2003RealTimeSL,Shotton:2013:RHP:2398356.2398381}. This technology allows to extract the position of the key joints in the body, which is suitable for 3D HAR problems. \textit{E.g.}, the latest version of the Microsoft Kinect $^{\text{TM}}$sensor \cite{Cruz2012KinectAR,Han2013EnhancedCV} (Kinect v2 sensor) can track the main 25 joints of human body in real-time speed as shown in Figure~\ref{fig:kinect}.
\\
\hspace*{0.3cm} In this paper, instead of feeding raw 3D joint coordinates directly to RNN-LSTMs as in many previous works \cite{7298714,zhu2016co,Liu2016SpatioTemporalLW,Shahroudy2016NTURA}, we propose the use of D-CNNs for learning motion features from skeleton sequences. The first step is to find a skeleton-based representation that can effectively capture the spatio-temporal evolutions of skeletons and that can also be easily learned by learning method as D-CNNs. One solution for this problem is to transform each skeleton sequence into a single color image (\textit{e.g.}, RGB image) in which the pixel values must have the ability to represent the movement of skeletons.\\[0.2cm]
\hspace*{0.3cm} To this end, we transform the 3D joint coordinates of each skeleton into a new space by normalizing their coordinates using a transformation function. Specifically, given a skeleton sequence $\mathcal{S}$ with \textbf{N} frames $[\textbf{F}_1, \textbf{F}_2,..., \textbf{F}_N]$, let $(x_i,y_i,z_i)$ be the 3D coordinates of each joint in frame $\{\textbf{F}_n\} \in \mathcal{S}, n \in [1, N]$. A normalization function $\mathcal{F}(\cdot)$ is used to transform all 3D joint coordinates to the range of [\textbf{0}, \textbf{255}] as follows:
\begin{equation}
(x'_i,y'_i,z'_i) = \mathcal{F}(x_i,y_i,z_i) 
\end{equation} 
\begin{equation}
x'_i = 255 \times \dfrac{(x_i - \min\{\mathcal{C}\})}{\max\{\mathcal{C}\} - \min\{\mathcal{C}\} } 
\end{equation} 
\begin{equation}
y'_i = 255 \times \dfrac{(y_i - \min\{\mathcal{C}\})}{\max\{\mathcal{C}\} - \min\{\mathcal{C}\} }
\end{equation}
\begin{equation}
z'_i = 255 \times \dfrac{(z_i - \min\{\mathcal{C}\})}{\max\{\mathcal{C}\} - \min\{\mathcal{C}\} }
\end{equation} 
where $(x'_i,y'_i,z'_i)$ is 3D joint coordinate in the normalized space $\mathcal{S}^{'}$. The $\max\{\mathcal{C}\}$ and $\min\{\mathcal{C}\}$ are the maximum and minimum values of all coordinates, respectively. To preserve the spatio-temporal information of skeleton movements, we stack all normalized frames according to the temporal order $\mathcal{S}^{'} = [\textbf{F}_1^{'},\textbf{F}_2^{'},...,\textbf{F}_N^{'}]$ to represent the whole action sequence. These elements are quantified to RGB color space and can be stored as RGB images. In this way, we convert the skeletal data to 3D tensors that will can then be fed into deep learning networks as the input for feature leaning and classification. Figure~\ref{fig:skeleton-rgb} illustrates the skeleton-to-image transformation process. \\[0.2cm]
\begin{figure}[h]
\centering
\includegraphics[width=8.5cm,height=4.5cm]{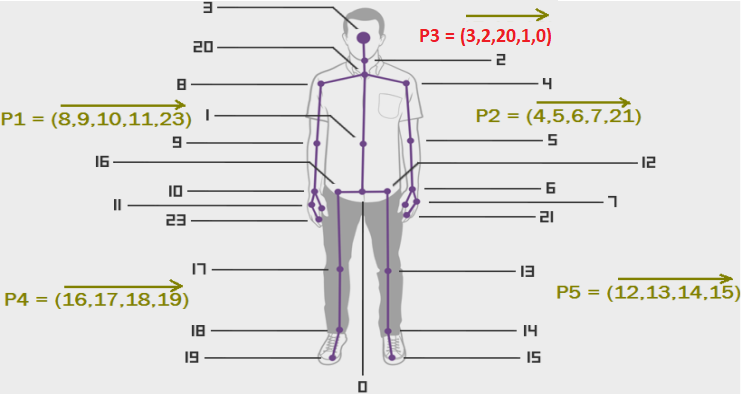} 
\caption{\label{fig:deviding-skeleton} Map of joints in each part from $\textbf{P}_1$ to $\textbf{P}_5$ on a skeleton provided by Microsoft Kinect v2 $^{\text{TM}}$sensor \cite{instance1290}.}
\end{figure}
\hspace*{0.3cm} Naturally, the human body is structured as four limbs and one trunk (see Figure~\ref{fig:kinect}). Simple actions can be performed through the movement of a limb, \textit{e.g.}, hand-waving, kicking forward, etc. More complex actions combine the movements of a group of limbs or the whole body, \textit{e.g.}, running or swimming. To keep the local characteristics of the human action \cite{yacoob1999parameterized,chaudhry2013bio}, we divide each skeleton into five parts, including two arms ($\textbf{P}_1$, $\textbf{P}_2$), two legs ($\textbf{P}_4$, $\textbf{P}_5$), and one trunk ($\textbf{P}_3$). In each part, the joints are concatenated according their physical connections as shown in Figure~\ref{fig:deviding-skeleton}. These parts are then connected in a sequential order:  $\textbf{P}_1 \to \textbf{P}_2 \to \textbf{P}_3 \to \textbf{P}_4 \to \textbf{P}_5$. The whole process of rearranging all frames in a sequence can be done by rearranging the order of rows of pixels in the color-based representation. This process is illustrated in Figure~\ref{fig:rearrange}.
\\
\\
Like that, we have encoded skeleton sequences into RGB images. Figure~\ref{fig:image-representation} shows some examples of RGB images obtained from input sequences of MSR Action3D dataset \cite{Li2010ActionRB}. These images will be learned and classified by D-CNN models. This way, the original skeleton sequence will be recognized via the corresponding image.  

\subsection{Deep residual networks for skeleton-based human action recognition} \label{residual-networks}

In this section, we introduce our network designs for recognizing human action from image-based representation obtained in Section~\ref{image-based representation}. Before that, in order to put our method into context, we will briefly review the key idea of ResNet \cite{He2016DeepRL}.
\subsubsection{Residual learning method}
\begin{figure}[H]
\centering
\includegraphics[width=8.6cm,height=6cm]{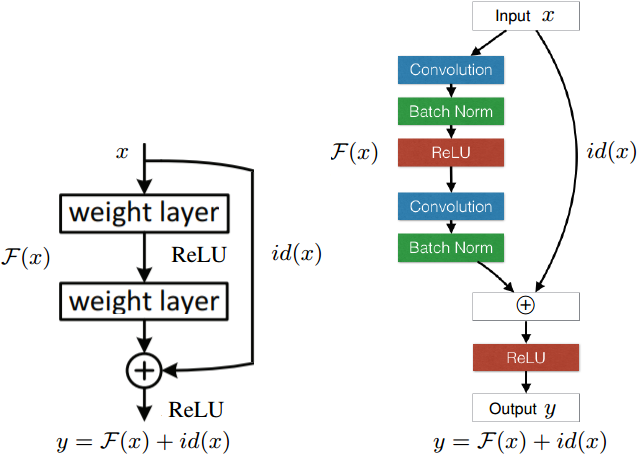}  \\
\caption{\label{fig:building-block} A building block of ResNet: (\textbf{a}) Information flow in a ResNet building block; (\textbf{b}) Implementation details of a ResNet block introduced in the original paper \cite{He2016DeepRL}. The symbol \textcircled{+} denotes element-wise addition.}
\label{Resnet-building-block}
\end{figure}

Very deep neural networks, especially D-CNNs have demonstrated to have a high performance on many visual-related tasks \cite{DBLP:journals/corr/SimonyanZ14a,Szegedy2015GoingDW}. However, D-CNNs are very difficult to optimize due to the vanishing gradients problem. If the network is deep enough, the error signal from the output layer can be completely attenuated on its way back to the input layer. In addition, another problem called ``the degradation phenomenon \cite{He2015ConvolutionalNN}'' also impedes the convergence of deeper networks. More specially, adding more layers to a deep network can lead to higher training/testing error. To overcome these challenges, ResNet \cite{He2016DeepRL} has been introduced. The key idea behind ResNet architectures is the presence of shortcut connections between input and output of each ResNet building block. These shortcut connections are implemented by identity mappings and are able to provide a path for gradients to back propagate to early layers in the network. This idea improves the information flow in ResNets and helps them to learn faster. Not only that, experimental results on standard datasets such as CIFAR-10 \cite{Krizhevsky2009LearningML}, and ImageNet \cite{ILSVRC15} for image classification tasks showed that the use of the shortcut connections in ResNet architectures can boost recognition performance. Figure~\ref{Resnet-building-block} shows information flow in a ResNet building block and their implementation details.
\hspace*{0.3cm} Mathematically, a layer or a series of layers in a typical CNN model will try to learn a mapping function  $y = \mathcal{F}(x)$ from input feature $x$. However, a ResNet building block will learn the function $y = \mathcal{F}(x) + id(x)$ where $id(x)$ is a identity function $id(x) = x$. The additional element $id(x)$ is the key factor that improves the information flow through layers during training ResNets. 
 
\subsubsection{Network design}
\begin{figure}[h]
\centering
\includegraphics[width=6cm,height=7.5cm]{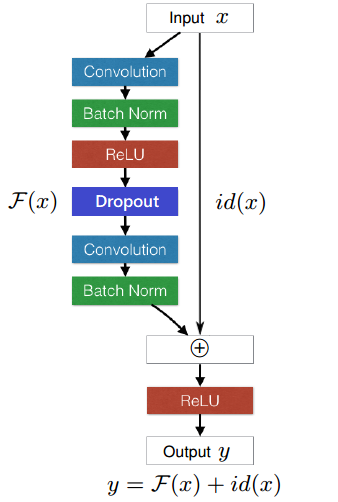} \\
\caption{ The proposed ResNet building block. Here, the symbol \textcircled{+} denotes element-wise addition.}
\label{our-ResNet-design}
\end{figure}

A deep ResNet can be constructed from multiple ResNet building blocks that are serially connected to each other. To search for the best recognition performance, we suggest different configurations of ResNet with 20, 32, 44, 56, and 110 layers. For further details, see Appendix. These networks are denoted as ResNet-20, ResNet-32, ResNet-44, ResNet-56, and ResNet-110, respectively. In our implementations, a ResNet building block performs the learning of a function $y = \mathcal{F}(x) + id(x)$ where $id(x) = x$ and $\mathcal{F}(x)$ is implemented by a sequence of layers: \textbf{Conv-BN-ReLU-Dropout-Conv-BN}. Specifically, each ResNet block uses the convolutional layers (\textbf{Conv}) with $\textbf{3} \times \textbf{3}$ filters. The Batch Normalization (\textbf{BN}) \cite{Ioffe2015BatchNA} and non-linear activation function \textbf{ReLU} \cite{Nair2010RectifiedLU} are applied after each \textbf{Conv}. To reduce possible overfitting, we add a Dropout layer \cite{Srivastava2014DropoutAS} with a rate of \textbf{0.5} into each ResNet block, located between two \textbf{Conv} layers and after \textbf{ReLU}. Finally, another \textbf{ReLU} layer \cite{Nair2010RectifiedLU} is used after the element-wise addition. The proposed ResNet unit is shown in Figure~\ref{our-ResNet-design}. We refer the interested reader to the Supplementary Materials to see details of the proposed network architectures. 
\\[0.2cm]
\hspace*{0.3cm} In this learning framework, all networks are designed for accepting images with size $\textbf{32} \times \textbf{32}$ pixels and trained in an end-to-end manner using Stochastic Gradient Descent (SGD) algorithm \cite{Bottou2010} from scratch. The last fully-connected layer of the network represents the action class scores and its size can be changed corresponding to the number of action classes.
\subsubsection{Learning to recognize actions from skeleton-based representations}
In order to recognize an action from a given skeleton sequence $\mathcal{S}$ with \textit{N} frames $[\textbf{F}_1, \textbf{F}_2,..., \textbf{F}_N]$,  we firstly encode all frames of $\mathcal{S}$ into a single RGB image $I_{\textbf{RGB}}$ as mentioned in section~\ref{image-based representation} 
\begin{equation}
I_{\textbf{RGB}} = [\textbf{F}_1^{'},\textbf{F}_2^{'},...,\textbf{F}_N^{'}], 
\end{equation}
where all elements of $\textbf{F}_{i | i \in [1, N]}^{'}$ were normalized to the range of $[\textbf{0}, \textbf{255}]$ as color pixels and rearranged according to their physical structure (see Figure~\ref{fig:rearrange}). We then propose the use of ResNets to learn and classify the obtained color images. During the training phase, we minimize the cross-entropy loss function $\mathcal{L}_{\mathcal{X}}(\textbf{y},\hat{\textbf{y}})$ between the action labels $\textbf{y}$ of $I_{\textbf{RGB}}$ and the predicted action labels $\hat{\textbf{y}}$ by the network over the training set $\mathcal{X}$. In other words, the network will be trained to solve the following problem
\begin{equation}
\textrm{\textbf{Arg}} \hspace{0.1cm} \underset{\mathcal{W}}{\text{min}}(\mathcal{L}_{\mathcal{X}}(\textbf{y},\hat{\textbf{y}})) = \textrm{\textbf{Arg}} \hspace{0.1cm} \underset{\mathcal{W}}{\text{min}} \left (   - \frac{1}{M} \sum_{i = 1}^{M}   \sum_{j = 1}^{C} \textbf{y}_{ij} \log \hat{\textbf{y}}_{ij}\right ),
\end{equation}
where $\mathcal{W}$ is the set of weights that will be learned by the network, $M$ denotes the number of samples in training set $\mathcal{X}$ and $C$ is the number of action classes.
\section{Experiments}
\label{sect:4}
\label{sec:experiment}
In this section, we evaluate the effectiveness of the proposed method on two benchmark datasets: MSR Action3D \cite{Li2010ActionRB} and NTU-RGB+D \cite{Shahroudy2016NTURA}. To have a fair comparison with state-of-the-art approaches in the literature, we follow the evaluation protocols as provided in the original papers \cite{Li2010ActionRB,Shahroudy2016NTURA}. Recognition performance will be measured by average classification accuracy.
\subsection{Datasets and experimental protocols}
\label{subsec:dataset}
\subsubsection{MSR Action3D Dataset \cite{Li2010ActionRB}}
This dataset\footnote{The MSR Action3D dataset can be obtained at: \url{https://www.uow.edu.au/\~wanqing/\#Datasets}} was collected with the Microsoft Kinect v1 $^{\text{TM}}$sensor. It contains 20 different actions. Each action was performed by 10 subjects for two or three times. The experiments were conducted on 557 valid sequences of the dataset after removing the defective sequences. We followed the same experimental protocol as many other authors, in which the whole data is divided into three subsets named as \textbf{AS1}, \textbf{AS2}, and \textbf{AS3}. For each subset, five subjects (with \textbf{ID}s: \textbf{1}, \textbf{3}, \textbf{5}, \textbf{7}, \textbf{9}) are selected for training and the rest (with \textbf{ID}s: \textbf{2}, \textbf{4}, \textbf{6}, \textbf{8}, \textbf{10}) for test. More details about the action classes in each subset can be found in Table~\ref{tab:1}.
\begin{table}[h]
\centering
\begin{tabular}{lll}
\hline
\textbf{AS1} & \textbf{AS2 }& \textbf{AS3} \\
\hline
\textit{Horizontal arm wave} & \textit{High arm wave} & \textit{High throw}\\ 
\textit{Hammer} & \textit{Hand catch} & \textit{Forward kick}\\
\textit{Forward punch} & \textit{Draw} x & \textit{Side kick}\\
\textit{High throw} &  \textit{Draw tick} & \textit{Jogging}\\
\textit{Hand clap} & \textit{Draw circle} & \textit{Tennis swing}\\
\textit{Bend} & \textit{Two hand wave} & \textit{Tennis serve}\\
\textit{Tennis serve} & \textit{Forward kick}& \textit{Golf swing}\\
\textit{Pickup} \& \textit{Throw} &  \textit{Side-boxing} & \textit{Pickup} \& \textit{Throw}\\ 
\hline
\end{tabular}
\caption{\label{tab:1} Three subsets \textbf{AS1}, \textbf{AS2}, and \textbf{AS3} of the MSR Action3D dataset \cite{Li2010ActionRB}.}
\end{table}
\subsubsection{NTU-RGB+D Dataset \cite{Shahroudy2016NTURA}}

The NTU-RGB+D dataset\footnote{Obtained at: \url{http://rose1.ntu.edu.sg/Datasets/actionRecognition.asp}} is a very large-scale RGB+D dataset. To the best of our knowledge, the NTU-RGB+D is currently the largest dataset that contains skeletal data for HAR. It is a very challenging dataset due to the large intra-class variations and multiple viewpoints. Specifically, the NTU-RGB+D provides more than 56,000 videos, collected from 40 subjects for 60 action classes. The list of action classes in NTU RGB+D dataset includes: \textit{drinking, eating, brushing teeth, brushing hair, dropping, picking up, throwing, sitting down, standing up (from sitting position), clapping, reading, writing, tearing up paper, wearing jacket, taking off jacket, wearing a shoe, taking off a shoe, wearing on glasses, taking off glasses, putting on a hat/cap, taking off a hat/cap, cheering up, hand waving, kicking something, reaching into self pocket, hopping, jumping up, making/answering a phone call, playing with phone, typing, pointing to something, taking selfie, checking time (on watch), rubbing two hands together, bowing, shaking head, wiping face, saluting, putting palms together, crossing hands in front, sneezing/coughing, staggering, falling down, touching head (headache), touching chest (stomachache/heart pain), touching back (back-pain), touching neck (neck-ache), vomiting, fanning self, punching/slapping other person, kicking other person, pushing other person, patting other's back, pointing to the other person, hugging, giving something to other person, touching other person's pocket, handshaking, walking towards each other, and walking apart from each other}. \\[0.2cm]
\hspace*{0.3cm} The NTU RGB+D data was collected by the Microsoft Kinect v2 $^{\text{TM}}$sensor. Each skeleton frame provides the 3D coordinates of 25 body joints. Two different evaluation criteria have been suggested, including \textbf{Cross-Subject} and \textbf{Cross-View}. In the \textbf{Cross-Subject} settings, 20 subjects (with \textbf{ID}s: \textbf{1}, \textbf{2}, \textbf{4}, \textbf{5}, \textbf{8}, \textbf{9}, \textbf{13}, \textbf{14}, \textbf{15}, \textbf{16}, \textbf{17}, \textbf{18}, \textbf{19}, \textbf{25}, \textbf{27}, \textbf{28}, \textbf{31}, \textbf{34}, \textbf{35}, and \textbf{38}) are used for training and the rest are used for test. For \textbf{Cross-View} settings, the sequences provided by cameras \textbf{2} and \textbf{3} are used for training and sequences provided by camera \textbf{1} are used for test.

\subsection{Data augmentation}  
\begin{figure}[h]
\centering
\includegraphics[width=8.5cm,height=2.5cm]{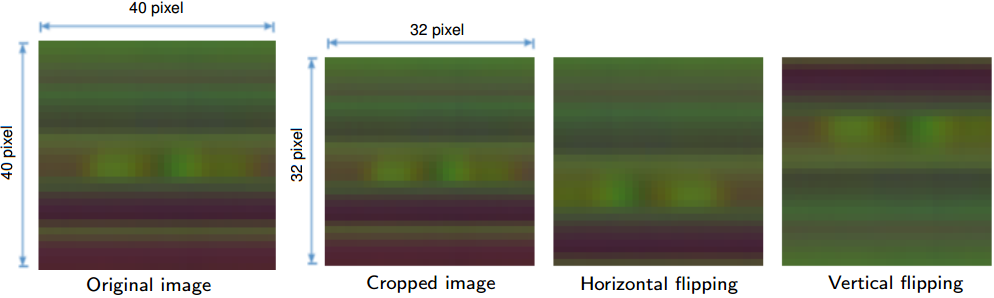} \\
\caption{\label{fig:data-augementation} Illustration of data augmentation methods applied on MSR Action3D dataset \cite{Li2010ActionRB}.}
\end{figure}

Very deep neural networks as ResNets \cite{He2016DeepRL} require a lot of data to optimize. D-CNNs cannot learn and classify images effectively when training with an insufficient data amount. Unfortunately, we have only \textbf{557} skeleton sequences (less than \textbf{30} images per class) on MSR Action3D dataset \cite{Li2010ActionRB}. Therefore, data augmentation has been applied to prevent ResNets from overfitting. We used random cropping, flip horizontally and vertically techniques to generate more training samples. More specifically, \textbf{8} $\times$ cropping has been applied on $\textbf{40}$  $\times$ $\textbf{40}$ images to create $\textbf{32}$ $\times$ $\textbf{32}$ images. Then, their horizontally and vertically flipped images are also created. Figure~\ref{fig:data-augementation} illustrates some data augmentation techniques on an image sample obtained from the MSR Action3D dataset \cite{Li2010ActionRB}. For the NTU-RGB+D dataset \cite{Shahroudy2016NTURA}, we do not apply any data augmentation method due to its very large-scale. 
\subsection{Implementation details} 
The image-based representations are computed directly from the original skeletons without any data preprocessing technique, \textit{e.g.}, transforming the coordinate system, using a fixed number of frames, or removing noise. In this project, the MatConvNet toolbox \cite{Vedaldi2015MatConvNetC} was used to design and train our deep learning networks. During the training phase, we use mini-batches of \textbf{128} images for ResNet-20, ResNet-32, ResNet-44, and ResNet-56 networks. For ResNet-110 network, we use mini-batches of \textbf{64} images. We initialize the weights randomly and train all networks for \textbf{120} epochs on a computer using Geforce GTX 1080 Ti GPU with 11GB RAM from scratch. The initial learning rate is set to \textbf{0.01} and is decreased to \textbf{0.001} at epoch \textbf{75}. The last \textbf{45} epochs use a learning rate of \textbf{0.0001}. The weight decay is set at \textbf{0.0001} and the momentum is \textbf{0.9}.
\begin{figure*}[h]
\centering
\includegraphics[width=5.8cm,height=4cm]{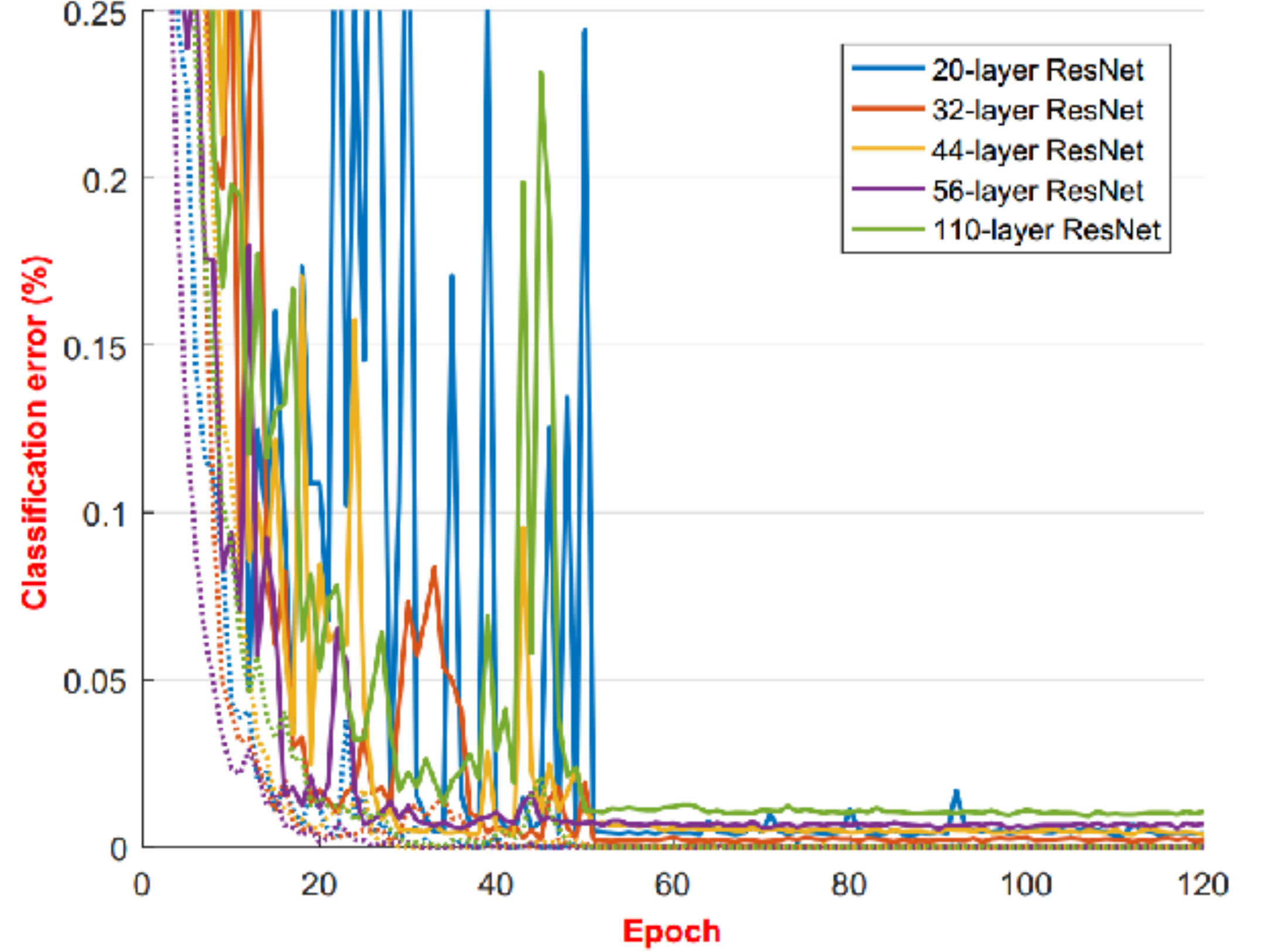}
\includegraphics[width=5.8cm,height=4cm]{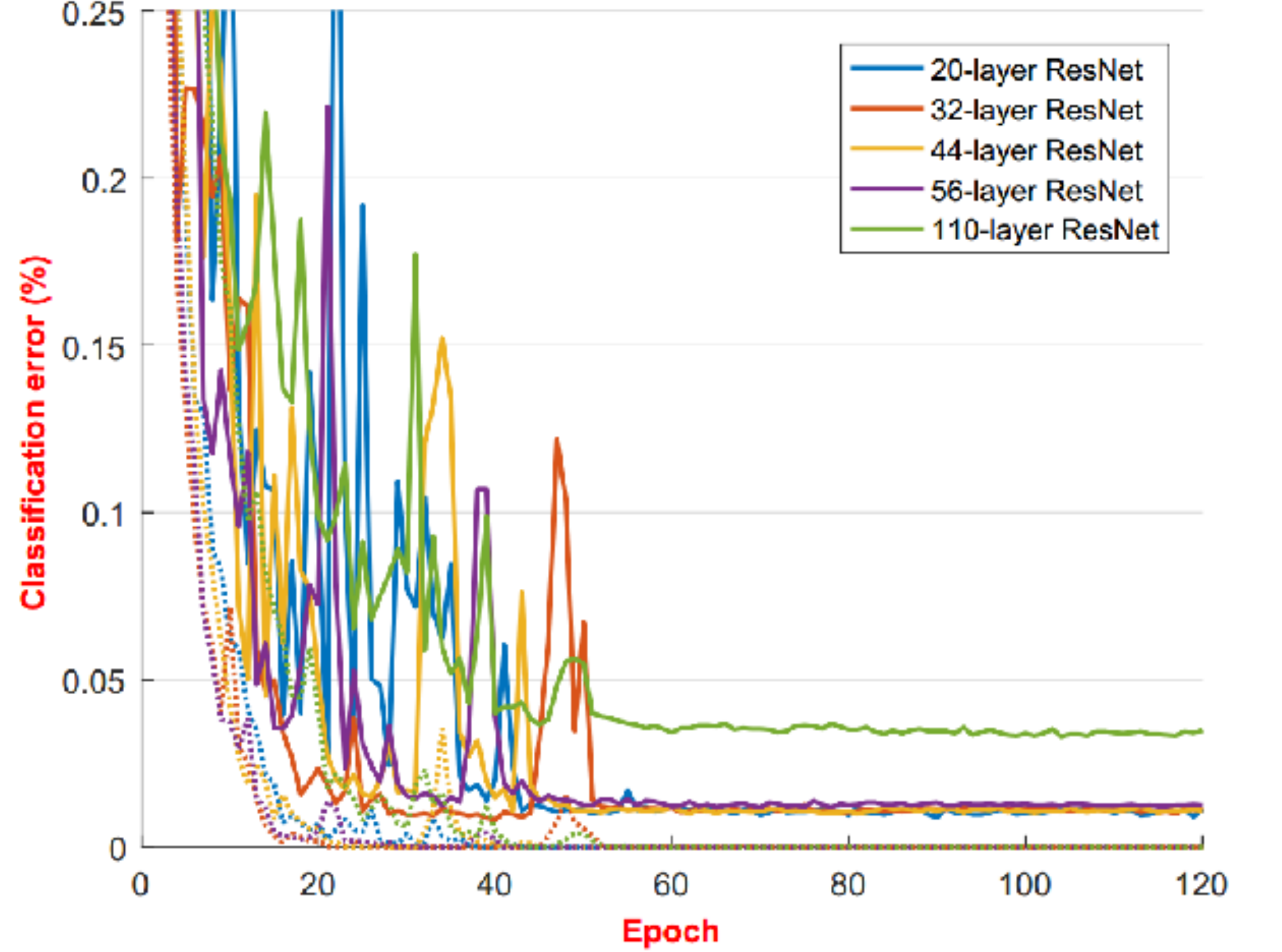}
\includegraphics[width=5.8cm,height=4cm]{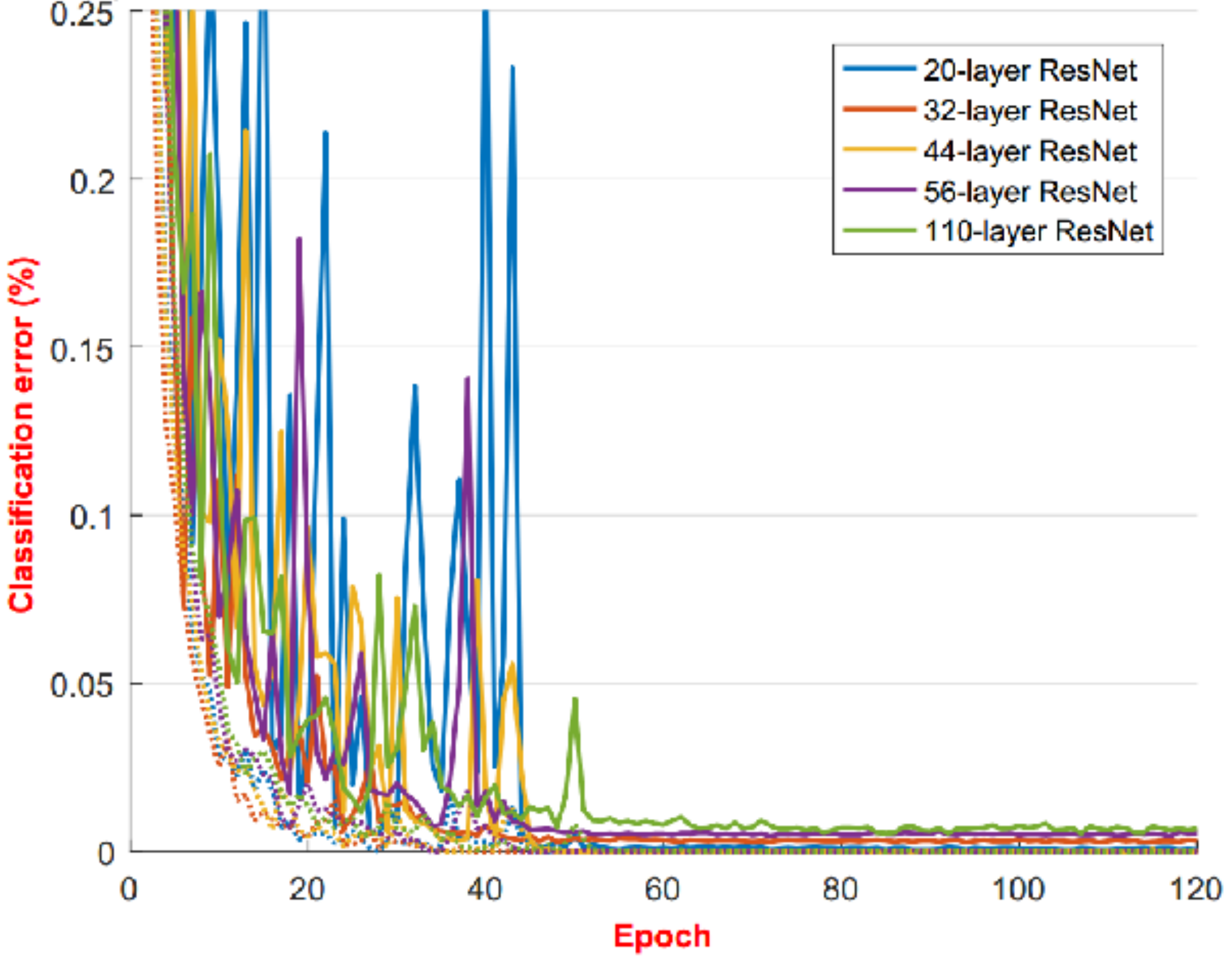} \\
\hspace*{0.2cm} \footnotesize{(a) Learning on AS1 subset} \hspace{2.6cm} \footnotesize{(b) Learning on AS2 subset} \hspace{2.6cm} \footnotesize{(c) Learning on AS3 subset} \\
\includegraphics[width=6cm,height=4cm]{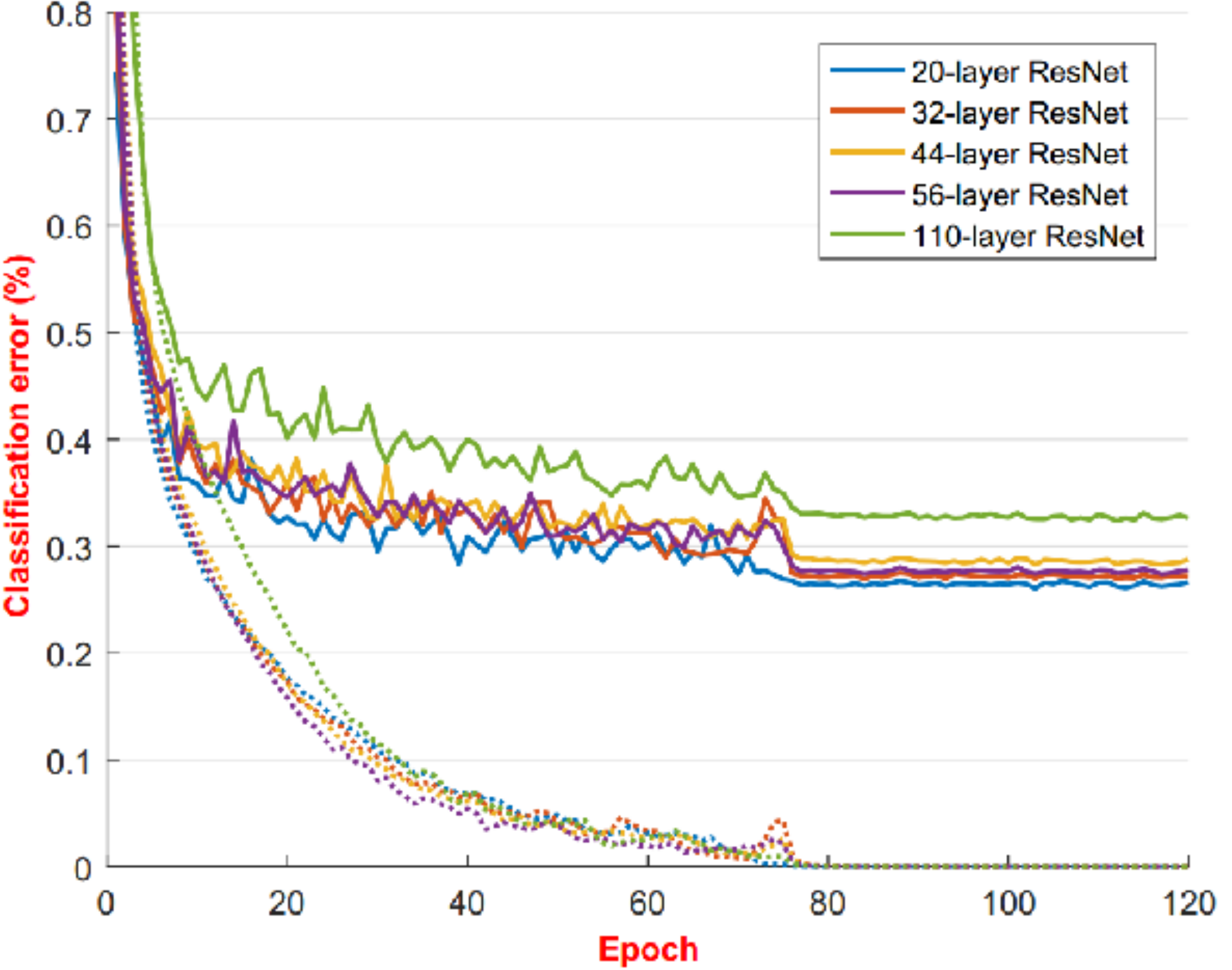} \hspace{2cm}
\includegraphics[width=6cm,height=4cm]{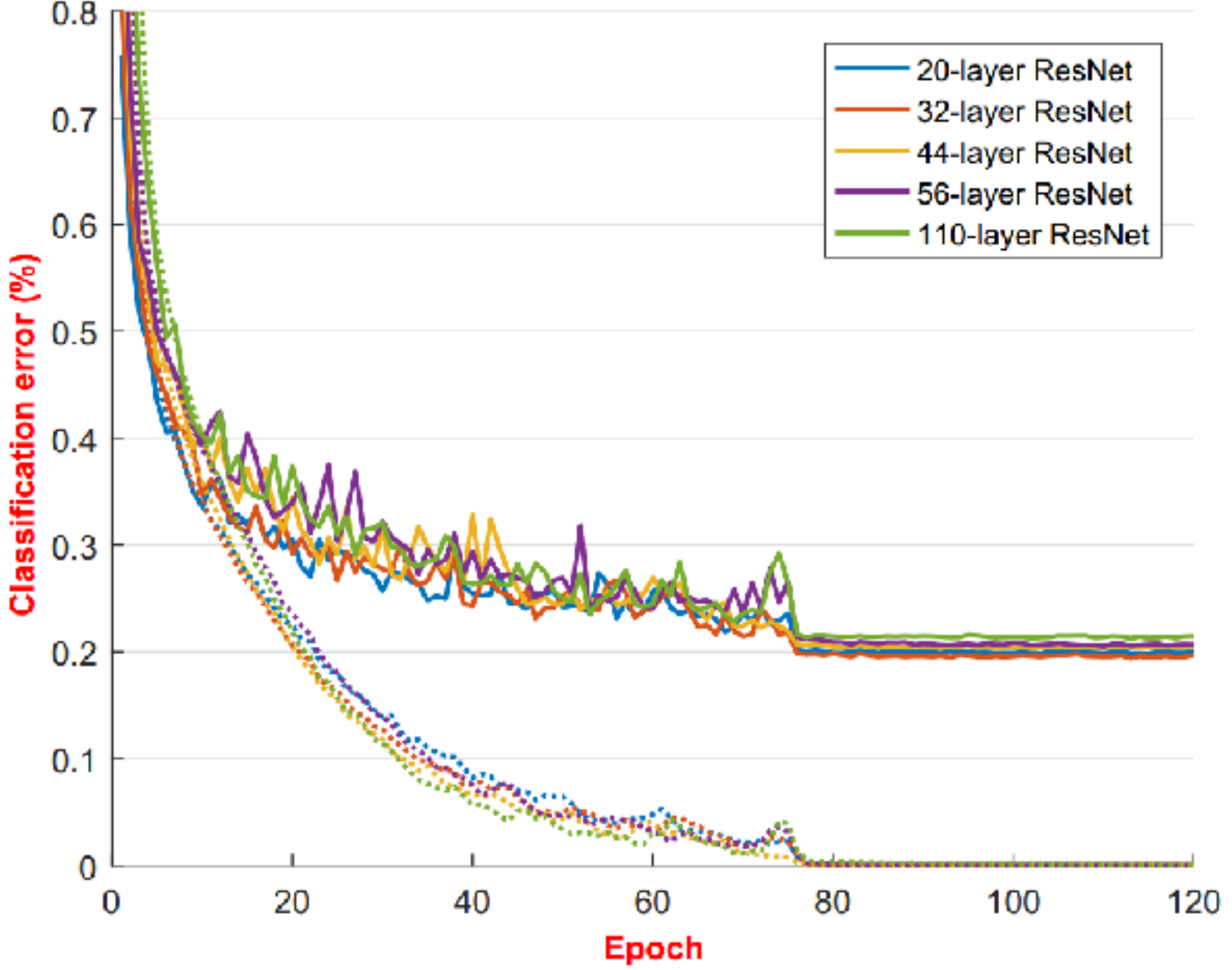}\\
\footnotesize{(d) Learning on NTU-RGD+D dataset (View-Cross Settings)}  \hspace{0.5cm} \footnotesize{(e) Learning on NTU-RGD+D dataset (Cross-Subject Settings)}
\caption{\label{curve1} Learning curves of 20-layer, 32-layer, 44-layer, 56 layer, and 110-layer ResNets on the MRS Action3D \cite{Li2010ActionRB} and the NTU-RGB+D \cite{Shahroudy2016NTURA} datasets. Dashed lines denote training errors while bold lines denote test errors. Best viewed on a computer, with zoomed-in and in color.}
\end{figure*}
\section{Experimental Results and Analysis}
\label{sect:5}
This section reports our experimental results on the MSR Action3D dataset \cite{Li2010ActionRB} and the NTU-RGB+D dataset \cite{Shahroudy2016NTURA}. We compare the obtained classification rates with state-of-the-art approaches in the literature \cite{Li2010ActionRB,vieira2012stop,xia2012view,Chaaraoui2013FusionOS,Chen2013RealtimeHA,Luo2013GroupSA,Gowayyed2013HistogramOO,Hussein:2013:HAR:2540128.2540483,qin2013gesture,Liang2013ThreeDM,Evangelidis2014SkeletalQH,Theodorakopoulos2014PosebasedHA,Gao2014HumanAR,Vieira2014OnTI,Chen2015ActionRF,7298714,Xu2015SpatioTemporalPM,jin2017action,Oreifej2013HON4DHO,Yang2014SuperNV,OhnBar2013JointAS,Evangelidis2014SkeletalQH,Misra2016ShuffleAL,Cippitelli2016EvaluationOA,Vemulapalli2014HumanAR,Vemulapalli2016RollingRF,7298714,Shahroudy2016NTURA,Luo2017UnsupervisedLO,Liu2016SpatioTemporalLW}. In addition, we also provide a detailed analysis about the computational efficiency of the proposed learning framework.
\subsection{Result on MSR Action3D dataset}
Experimental results on MSR Action3D dataset \cite{Li2010ActionRB} are shown in Table~\ref{tab_result_MSR}. We achieved the best classification accuracy with the \textbf{ResNet-32} model. More specifically, classification accuracies are \textbf{99.80\%} on \textbf{AS1}, \textbf{99.00\%} on \textbf{AS2}, and \textbf{99.80\%} on \textbf{AS3}. We obtained a total average accuracy of \textbf{99.53\%}. Our result outperforms many previous works \cite{Li2010ActionRB,vieira2012stop,xia2012view,Chaaraoui2013FusionOS,Chen2013RealtimeHA,Luo2013GroupSA,Gowayyed2013HistogramOO,Hussein:2013:HAR:2540128.2540483,qin2013gesture,Liang2013ThreeDM,Evangelidis2014SkeletalQH,Theodorakopoulos2014PosebasedHA,Gao2014HumanAR,Vieira2014OnTI,Chen2015ActionRF,7298714,Xu2015SpatioTemporalPM,jin2017action,10.1007/978-3-319-77383-4_35,Theodorakopoulos2014PosebasedHA,7298714,Chen2015ActionRF,Xu2015SpatioTemporalPM,jin2017action,luvizon2017learning,Luo2013GroupSA} (see Table~\ref{tab:4}). Moreover, this result also indicates that the proposed ResNet architecture learns image features better than the original ResNet architecture \cite{iet:/content/conferences/10.1049/cp.2017.0154}.   The learning curves of all proposed ResNets on \textbf{AS1}, \textbf{AS2}, \textbf{AS3}  are shown in Figure~\ref{curve1}a, Figure~\ref{curve1}b, and Figure~\ref{curve1}c, respectively.
\begin{table}[h]
\centering
\begin{tabular}{lllll}
\hline
\hspace*{0.03cm} \textbf{Network} & \hspace*{0.03cm} \textbf{AS1} & \hspace*{0.03cm} \textbf{AS2}& \hspace*{0.03cm} \textbf{AS3} & \hspace*{0.03cm} \textbf{Aver.} \\
\hline
\hspace*{0.025cm} ResNet-20  & \hspace*{0.03cm} 99.60\% & \hspace*{0.03cm} 98.90\% & \hspace*{0.03cm} 100.0\% & \hspace*{0.03cm} 99.50\% \\
\cellcolor{gray!50} \textbf{ResNet-32}  & \cellcolor{gray!50} \textbf{99.80}\% & \cellcolor{gray!50} \textbf{99.00}\% & \cellcolor{gray!50} \textbf{99.80}\% &  \cellcolor{gray!50} \textbf{99.53}\% \\
\hspace*{0.025cm} ResNet-44  & \hspace*{0.03cm} 99.50\% & \hspace*{0.03cm} 98.90\% & \hspace*{0.03cm} 100.0\% & \hspace*{0.03cm} 99.48\% \\
\cellcolor{gray!50} ResNet-56  & \cellcolor{gray!50} 99.40\% & \cellcolor{gray!50} 98.70\% & \cellcolor{gray!50} 99.50\% & \cellcolor{gray!50} 98.20\% \\
\hspace*{0.025cm} ResNet-110 & \hspace*{0.03cm} 99.00\% & \hspace*{0.03cm} 96.50\% & \hspace*{0.03cm} 99.30\% & \hspace*{0.03cm} 99.28\% \\
\hline
\end{tabular}
\caption{\label{tab_result_MSR} Test accuracies (\%) of our proposed ResNets on \textbf{AS1}, \textbf{AS2}, and \\
 \textbf{AS3} subsets. The best results and configuration are highlighted in \textbf{bold}.}
\end{table}
\begin{table}[h]
\centering
\begin{tabular}{lllll}
 	\hline
\hspace*{0.03cm} {\footnotesize \textbf{Method}}  & {\footnotesize \hspace*{0.05cm} \textbf{AS1}} & {\footnotesize \hspace*{0.03cm} \textbf{AS2 }} & {\footnotesize \hspace*{0.03cm} \textbf{AS3}} & {\footnotesize \hspace*{0.03cm} \textbf{Aver.}} \\
	\hline
    \hspace*{0.03cm} {\scriptsize Bag of 3D Points \cite{Li2010ActionRB}}  & {\scriptsize \hspace*{0.03cm} 72.90\% } & {\scriptsize \hspace*{0.03cm} 71.90\% } & {\scriptsize \hspace*{0.03cm} 79.20\%} &  {\scriptsize \hspace*{0.03cm} 74.67\%} \\
    \hline
      \cellcolor{gray!50} {\scriptsize Motion Trail Model \cite{Liang2013ThreeDM}}  & {\scriptsize \cellcolor{gray!50} 73.70\%} & {\scriptsize \cellcolor{gray!50} 81.50\% } & {\scriptsize \cellcolor{gray!50} 81.60\% } & {\scriptsize \cellcolor{gray!50} 78.93\% } \\
    \hline
    \hspace*{0.03cm} {\scriptsize Histograms of 3D Joints \cite{xia2012view}}  & {\scriptsize \hspace*{0.03cm} 87.98\%} & {\scriptsize \hspace*{0.03cm} 85.48\% } & {\scriptsize \hspace*{0.03cm} 63.46\%} & {\scriptsize \hspace*{0.03cm} 78.97\%} \\
    \hline
     \cellcolor{gray!50} {\scriptsize Motion and Shape Features \cite{qin2013gesture}}  &  {\scriptsize \cellcolor{gray!50} 81.00\%} & {\scriptsize \cellcolor{gray!50} 79.00\%} &  {\scriptsize \cellcolor{gray!50} 82.00\%} & {\scriptsize  \cellcolor{gray!50} 80.66\%} \\
    \hline
    \hspace*{0.03cm} {\scriptsize Spatial-temporal Features  \cite{hbali2017skeleton}}   	& {\scriptsize \hspace*{0.03cm} 77.36\%} & {\scriptsize  \hspace*{0.03cm} 73.45\%} & {\scriptsize \hspace*{0.03cm} 91.96\%} & {\scriptsize \hspace*{0.03cm} 80.92\%} \\
	\hline
 \cellcolor{gray!50} {\scriptsize Space-time Occupancy \cite{vieira2012stop}}  & {\scriptsize	\cellcolor{gray!50} 84.70\%} & {\scriptsize \cellcolor{gray!50} 81.30\%} & {\scriptsize \cellcolor{gray!50} 88.40\%} & {\scriptsize \cellcolor{gray!50} 84.80\%} \\
    \hline
     \hspace*{0.03cm} {\scriptsize Improved Space-time Occ. \cite{Vieira2014OnTI}}  & 	{\scriptsize \hspace*{0.03cm} 91.70\%} & {\scriptsize \hspace*{0.03cm} 72.20\%} & {\scriptsize \hspace*{0.03cm} 98.60\%} & {\scriptsize \hspace*{0.03cm} 87.50\%} \\
      \hline
    \cellcolor{gray!50}   {\scriptsize HON4D \cite{Oreifej2013HON4DHO}}   	&  {\scriptsize \cellcolor{gray!50} N/A} &   {\scriptsize \cellcolor{gray!50} N/A} &  {\scriptsize \cellcolor{gray!50} N/A} &  {\scriptsize \cellcolor{gray!50} 88.89\%} \\
	\hline
    \hspace*{0.03cm} {\scriptsize Skeletal Quads \cite{Evangelidis2014SkeletalQH}}  & {\scriptsize \hspace*{0.03cm} 88.39\%} & {\scriptsize \hspace*{0.03cm} 86.61\%} & {\scriptsize \hspace*{0.03cm} 94.59\%} & {\scriptsize \hspace*{0.03cm} 89.86\%} \\
    \hline
    \cellcolor{gray!50} {\scriptsize Multi-modality Information \cite{Gao2014HumanAR}}  & {\scriptsize \cellcolor{gray!50} 92.00\%}	& {\scriptsize \cellcolor{gray!50} 85.00\%} &  {\scriptsize \cellcolor{gray!50} 93.00\%} & {\scriptsize \cellcolor{gray!50} 90.00\%} \\
      \hline
     \hspace*{0.03cm} {\scriptsize Depth Motion Maps \cite{Chen2013RealtimeHA}}  & {\scriptsize \hspace*{0.03cm} 96.20\%} &  {\scriptsize \hspace*{0.03cm} 83.20\%} & {\scriptsize \hspace*{0.03cm} 92.00\%} & {\scriptsize \hspace*{0.03cm} 90.47\%} \\
    \hline
     \cellcolor{gray!50}  {\scriptsize Covariance Descriptors \cite{Hussein:2013:HAR:2540128.2540483}}  & {\scriptsize \cellcolor{gray!50} 88.04\%} & {\scriptsize \cellcolor{gray!50} 89.29\%} & {\scriptsize \cellcolor{gray!50} 94.29\%} & {\scriptsize \cellcolor{gray!50} 90.53\%} \\ 
    \hline
     \hspace*{0.03cm} {\scriptsize HOD \cite{Gowayyed2013HistogramOO}}  &  {\scriptsize \hspace*{0.03cm} 92.39\%} & {\scriptsize \hspace*{0.03cm} 90.18\%} & {\scriptsize \hspace*{0.03cm} 91.43\%} & {\scriptsize \hspace*{0.03cm} 91.26\%} \\
    \hline
     \cellcolor{gray!50} {\scriptsize Moving Pose \cite{zanfir2013moving}}  &  {\scriptsize \cellcolor{gray!50} N/A } & {\scriptsize \cellcolor{gray!50} N/A} & {\scriptsize \cellcolor{gray!50} N/A} & {\scriptsize \cellcolor{gray!50} 91.70\%} \\
    \hline
     {\scriptsize \hspace*{0.03cm} Skeletal and Silhouette Fusion \cite{Chaaraoui2013FusionOS}}   	& {\scriptsize \hspace*{0.03cm} 92.38\%} & {\scriptsize  \hspace*{0.03cm} 86.61\% } & {\scriptsize \hspace*{0.03cm} 96.40\%} & {\scriptsize \hspace*{0.03cm} 91.80\% }\\
	\hline

     \cellcolor{gray!50} {\scriptsize Improved Key Poses \cite{10.1007/978-3-319-77383-4_35}}  & {\scriptsize \cellcolor{gray!50}  91.53\%} &{\scriptsize \cellcolor{gray!50} 90.23\%} & {\scriptsize \cellcolor{gray!50} 97.06\%} & {\scriptsize \cellcolor{gray!50} 92.94\%} \\
     \hline
   
     \hspace*{0.03cm} {\scriptsize Pose-based Representation \cite{Theodorakopoulos2014PosebasedHA}}  & {\scriptsize \hspace*{0.03cm} 91.23\%} & {\scriptsize \hspace*{0.03cm} 90.09\%} & {\scriptsize \hspace*{0.03cm} 99.50\%} & {\scriptsize \hspace*{0.03cm} 93.61\%} \\
     \hline

      \cellcolor{gray!50} {\scriptsize H-RNN \cite{7298714}}  & {\scriptsize \cellcolor{gray!50} 93.33\%} & {\scriptsize \cellcolor{gray!50} 94.64\%} &  {\scriptsize \cellcolor{gray!50} 95.50\%} & {\scriptsize \cellcolor{gray!50} 94.49\%} \\
       \hline
       \hspace*{0.03cm}   {\scriptsize Local Binary Patterns \cite{Chen2015ActionRF}}  &	{\scriptsize \hspace*{0.03cm} 98.10\%} &	{\scriptsize \hspace*{0.03cm} 92.00\%} &	{\scriptsize \hspace*{0.03cm} 94.60\%} &	{\scriptsize \hspace*{0.03cm} 94.90\%} \\
       \hline
      
       \cellcolor{gray!50} {\scriptsize Spatio-temporal Pyramid \cite{Xu2015SpatioTemporalPM}}  &	{\scriptsize \cellcolor{gray!50} 99.10\% } &	{\scriptsize \cellcolor{gray!50} 92.90\%}  &	{\scriptsize \cellcolor{gray!50} 96.40\% } &	{\scriptsize \cellcolor{gray!50} 96.10\%} \\
       \hline
       \hspace*{0.03cm} {\scriptsize Depth Motion Maps \cite{jin2017action}}  & {\scriptsize \hspace*{0.03cm}  99.10\%} & {\scriptsize \hspace*{0.03cm}  92.30\%} &  {\scriptsize \hspace*{0.03cm} 98.20\%} & {\scriptsize \hspace*{0.03cm} 96.50\%} \\
        \hline
         \cellcolor{gray!50} {\scriptsize Features Combination \cite{luvizon2017learning}}  &  \cellcolor{gray!50} {\scriptsize N/A } & \cellcolor{gray!50}  {\scriptsize  N/A} & \cellcolor{gray!50} {\scriptsize  N/A} &  {\scriptsize  \cellcolor{gray!50} 97.10\% }\\
    \hline
          \hspace*{0.03cm}  {\scriptsize Group Sparsity \cite{Luo2013GroupSA}}  &  \hspace*{0.03cm}{\scriptsize 97.20\% } & \hspace*{0.03cm}   {\scriptsize 95.50\%} & \hspace*{0.03cm}  {\scriptsize  99.10\%} & \hspace*{0.03cm}   {\scriptsize  97.26\% }\\
    \hline
       {\scriptsize \cellcolor{gray!50} \textbf{Our best configuration}}  &  \cellcolor{gray!50} {\scriptsize  \textbf{99.80\%}} & \cellcolor{gray!50} {\scriptsize  \textbf{99.00\%}} & \cellcolor{gray!50} {\scriptsize   \textbf{99.80\%}} & \cellcolor{gray!50} {\scriptsize  \textbf{99.53\%}} \\
\end{tabular}
\caption{ Comparison with the state-of-the-art approaches on MSR Action3D \\ dataset  \cite{Li2010ActionRB}. The best performances are in \textbf{bold}.}
\label{tab:4}
\end{table}
\subsection{Result on NTU-RGB+D dataset}
As shown in Table~\ref{tab:3}, the proposed learning framework reaches competitive results with an accuracy of \textbf{73.40}\% on \textbf{Cross-Subject} evaluation and an accuracy of \textbf{80.40}\% on \textbf{Cross-View} evaluation. The obtained results indicate that our method can deal with large intra-class variations and multiple viewpoints dataset as NTU-RGB+D \cite{Shahroudy2016NTURA}. Table~\ref{tab:5} provides a comparison with published studies \cite{Li2010ActionRB,Liang2013ThreeDM,xia2012view,qin2013gesture,hbali2017skeleton,vieira2012stop,Vieira2014OnTI,Oreifej2013HON4DHO,Evangelidis2014SkeletalQH,Gao2014HumanAR,Chen2013RealtimeHA,Hussein:2013:HAR:2540128.2540483,Gowayyed2013HistogramOO,Chaaraoui2013FusionOS}. It is clear that our method surpasses many previous approaches in the same experimental conditions. The learning curves of all ResNet configurations on two evaluation settings are shown in Figure~\ref{curve1}d and Figure~\ref{curve1}e. 
\begin{table}[h]
\label{tab_result}
\centering
\begin{tabular}{lll}
\hline
\hspace*{0.03cm} \textbf{Network} & \hspace*{0.1cm} \textbf{Cross-Subject} & \hspace*{0.08cm} \textbf{Cross-View} \\
\hline
\hspace*{0.03cm} ResNet-20  & \hspace*{0.53cm}  \textbf{73.40}\% & \hspace*{0.5cm} 80.10\%   \\
\cellcolor{gray!50} \textbf{ResNet-32}  & \hspace*{0.45cm} \cellcolor{gray!50} 73.10\% & \hspace*{0.43cm} \cellcolor{gray!50} \textbf{80.40}\%  \\
\hspace*{0.03cm} ResNet-44  & \hspace*{0.53cm} 72.40\% & \hspace*{0.5cm} 79.80\%  \\
\cellcolor{gray!50} ResNet-56  & \hspace*{0.45cm} \cellcolor{gray!50} 73.00\% & \hspace*{0.43cm} \cellcolor{gray!50} 79.80\%  \\
\hspace*{0.03cm} ResNet-110 & \hspace*{0.54cm} 67.40\% & \hspace*{0.5cm} 78.60\%  \\
\hline
\end{tabular}
\caption{Test accuracies (\%) of our proposed networks on Cross-Subject \\ and Cross-View settings \cite{Shahroudy2016NTURA}. The best results and configuration are in \textbf{bold}.}
\label{tab:3}
\end{table}
\\
\begin{table}[h]
\centering
\begin{tabular}{lll}
 	\hline
\hspace*{0.03cm} \textbf{Method}  & \hspace*{0.03cm} \textbf{Cross-Subject} & \hspace*{0.03cm} \textbf{Cross-View }\\
    \hline
\hspace*{0.03cm}  HON4D \cite{Oreifej2013HON4DHO}  & 	\hspace*{0.03cm} 30.56\% & \hspace*{0.03cm} 7.26\% \\
    \hline
 \cellcolor{gray!50} Super Normal Vector \cite{Yang2014SuperNV}  &  \cellcolor{gray!50} 31.82\% &   \cellcolor{gray!50} 13.61\% \\
    \hline
 \hspace*{0.03cm} Joint Angles + HOG2 \cite{OhnBar2013JointAS}   	&  \hspace*{0.03cm} 32.24\% &   \hspace*{0.03cm} 22.27\%  \\
	\hline
    \cellcolor{gray!50} Skeletal Quads \cite{Evangelidis2014SkeletalQH}   	&  \cellcolor{gray!50} 38.62\% &   \cellcolor{gray!50} 41.36\%  \\
    \hline
\hspace*{0.03cm} Shuffle and Learn  \cite{Misra2016ShuffleAL}  & \hspace*{0.03cm} 47.50\% & \hspace*{0.03cm} N/A  \\
    \hline
     \cellcolor{gray!50} Histograms of Key Poses \cite{Cippitelli2016EvaluationOA}  &  \cellcolor{gray!50} 48.90\% &   \cellcolor{gray!50} 57.70\%  \\
    \hline
    \hspace*{0.03cm}  Lie Group \cite{Vemulapalli2014HumanAR}  & \hspace*{0.03cm} 50.08\% & \hspace*{0.03cm} 52.76\%  \\
    \hline
    \cellcolor{gray!50} Rolling Rotations  \cite{Vemulapalli2016RollingRF}  &  \cellcolor{gray!50} 52.10\% & \cellcolor{gray!50} 53.40\%  \\
    \hline
    \hspace*{0.03cm} H-RNN \cite{7298714} (reported in \cite{Shahroudy2016NTURA})  & \hspace*{0.03cm} 59.07\% & \hspace*{0.03cm} 63.79\% \\
    \hline
    \cellcolor{gray!50} P-LSTM \cite{Shahroudy2016NTURA}  &  \cellcolor{gray!50} 62.93\% &  \cellcolor{gray!50} 70.27\% \\ 
    \hline
 \hspace*{0.03cm} Long-Term Motion \cite{Luo2017UnsupervisedLO}  & \hspace*{0.03cm} 66.22\% & \hspace*{0.03cm}  N/A \\ 
    \hline
    \cellcolor{gray!50} Spatio-temporal LSTM \cite{Liu2016SpatioTemporalLW}  & \cellcolor{gray!50} 69.20\% & \cellcolor{gray!50} 77.70\%\\ 
    \hline
       \hspace*{0.03cm} \textbf{Our best configuration}  & \hspace*{0.03cm} \textbf{73.40\%} & \hspace*{0.03cm} \textbf{80.40\%}\\ 
\end{tabular}
\caption{\label{tab:5} Comparison with state-of-the-art methods on NTU-RGB+D dataset \\ \cite{Shahroudy2016NTURA}. The best results and configuration are marked in \textbf{bold}. }
\end{table}
\subsection{Convergence rate analysis}
Figure~\ref{curve1} shows the learning curves of the five proposed deep learning networks on the MSR Action3D \cite{Li2010ActionRB} and the NTU-RGB+D \cite{Shahroudy2016NTURA} datasets. We can find that the training convergence rate of the networks are different on the two datasets. More specifically, on MSR Action3D dataset \cite{Li2010ActionRB}, the proposed networks exhibit rapid convergence after \textbf{50} epochs at the beginning of the training process. Meanwhile, these networks can only start to converge after near \textbf{80} epochs on the NTU-RGB+D dataset \cite{Shahroudy2016NTURA}. This phenomenon can be explained by the complexity of feature spaces generated by the two datasets. With more than \textbf{56,000} videos collected from \textbf{40} subjects for \textbf{60} action classes, it is clear that the NTU-RGB+D dataset \cite{Shahroudy2016NTURA} is more complex than the MSR Action3D dataset \cite{Li2010ActionRB}. This leads to deep learning networks needing more iterations to start convergence.

As shown in \textbf{Table 2} and \textbf{Table 4}, the best recognition accuracy and configuration are highlighted in \textbf{bold}. More specifically, our experimental result on the MSR Action3D indicated that the baseline ResNet-32 achieved the best overall accuracy (99.53\%). On the NTU RGB+D dataset, the ResNet-32 network was the best version on the Cross-View setting (80.40\%). However, it worked worse than the ResNet-20 on the Cross-Subject setting. This is a weak aspect 
of the proposed approach. This limitation could be overcome by using ensemble learning techniques \cite{dietterich2000ensemble}.

\subsection{Analysis of training and prediction time} \label{prediction-time}
\begin{figure}[h]
\centering
\includegraphics[width=8cm,height=5cm]{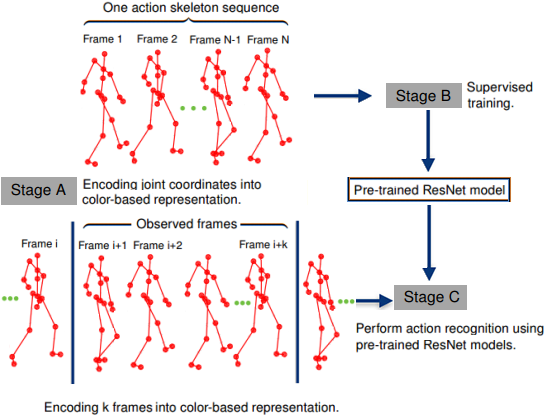}
\caption{\label{pipeline} Three main stages of the proposed deep learning framework.}
\end{figure}

We take the AS1 subset of the MSR Action3D dataset \cite{Li2010ActionRB} and the proposed ResNet-32 network for illustrating the computational efficiency of our deep learning framework. As shown in Figure~\ref{pipeline}, the proposed framework contains three main stages, including the encoding process from skeleton sequences into RGB images \textbf{(Stage A)},  the supervised training stage \textbf{(Stage B)}, and the prediction stage \textbf{(Stage C)}. With the implementation in Matlab using MatconvNet toolbox \cite{Vedaldi2015MatConvNetC}\footnote{MatconvNet is an open source library and can be downloaded at address: \url{http://www.vlfeat.org/matconvnet/}.} on a single NVIDIA GeForce GTX 1080 Ti GPU\footnote{More details about the GPU specification, please refer to: \url{https://www.nvidia.com/en-us/geforce/products/10series/geforce-gtx-1080-ti/}.}, without parallel processing, we take on average $\textbf{4.15} \times\textbf{10}^{-3}$s per sequence during training stage. While the prediction time, including the time for encoding skeletons into RGB images and classification by the pre-trained ResNet, takes on average $\textbf{21.84} \times\textbf{10}^{-3}$s per sequence (see Table~\ref{execution-time}). These results verify the effectiveness of our proposed method, not only in terms of accuracy but also in terms of computational efficiency.
\begin{table}[h]
\label{complexity}
\centering
\begin{tabular}{ll}
\hline
\hspace*{0.4cm} \textbf{Stage} &\hspace*{1cm} \textbf{Average processing time}  \\
\hline
\hline
\hspace*{0.4cm}  \textbf{Stage A} & \hspace*{1cm}  $\textbf{20.8} \times\textbf{10}^{-3}$s per sequence (CPU time) \\
\hline
\hspace*{0.4cm}  \textbf{Stage B} & \hspace*{1cm}  $\textbf{4.15} \times\textbf{10}^{-3}$s per sequence (GPU time) \\
\hline
\hspace*{0.4cm}  \textbf{Stage C} & \hspace*{1cm}  $\textbf{21.84} \times\textbf{10}^{-3}$s per sequence (CPU time)\\
\hline
\end{tabular}
\caption{\label{execution-time} Execution time of each component of the proposed learning\\ framework. }
\end{table}
\label{sect:6}
\section{Conclusion and future work}
\label{sect:7}
In this paper, we have proposed a new 3D motion representation and an end-to-end deep learning framework based on ResNets \cite{He2016DeepRL} for human action recognition from skeleton sequences. To this end, we transformed the 3D joint coordinates carried in skeleton sequences into RGB images via a color encoding process. We then designed and trained different D-CNNs based on the ResNet architecture for learning the spatial and temporal dynamics of human motion from image-coded representations. The experimental results on two well-established datasets, including the largest RGB-D dataset currently available, demonstrated that our method can achieve state-of-the-art performance whilst requiring a low computation cost for the training and prediction stages.
For future work, we aim to extend this research by investigating a new skeleton encoding method, in which the Euclidean distances and orientation relations between joints will be exploited. In addition, we plan to build a better feature learning and classification framework by experimenting some new and potential architectures based on the idea of ResNet such as Inception Residual Networks (Inception-ResNets) \cite{article_inception_v4} or Densely Connected Convolutional Networks (DenseNet \cite{Huang2017DenselyCC}). In order to overcome the limitations of Kinect sensors when dealing with HAR in which the lighting conditions are available, we will also consider some new approaches for estimating human body key-points such as deep learning based approaches \cite{cao2017realtime,Guler2018DensePose}, KWE and KCE methods \cite{akbarizadeh2012new}. We hope that our study will open a new door for the computer vision community on exploiting the potentials of very deep convolutional neural networks and skeletal data for 3D human action recognition.\\

\section*{Acknowledgment}

This research was carried out at the Cerema Research Center (CEREMA) and Toulouse Institute of Computer Science Research (IRIT), Toulouse, France. Sergio A. Velastin is grateful for funding received from the Universidad Carlos III de Madrid, the European Union's Seventh Framework Programme for Research, Technological Development and demonstration under grant agreement N. 600371, el Ministerio de Economia, Industria y Competitividad (COFUND2013-51509) el Ministerio de Educaci\'on, cultura y Deporte (CEI-15-17) and Banco Santander.

\bibliographystyle{ietConference}
\bibliography{reference}
\noindent \\
\textbf{Appendix}\\
\\
This section describes the network architectures in detail. To build 20-layer, 32-layer, 44-layer, 56-layer, and 110-layer networks, we stack the proposed ResNet building units as following:\\
\\
\begin{tabular}{p{8.6cm}}
\hspace*{2cm} {\small \textbf{Baseline 20-layer ResNet architecture}} \\
\hline
\hline
{\small 3x3 Conv., 16 filters, BN, ReLU}\\
{\small Residual unit: Conv.-BN-ReLU-Dropout-Conv.,16 filters-\textcircled{+}-ReLU }\\
{\small Residual unit: Conv.-BN-ReLU-Dropout-Conv.,16 filters-\textcircled{+}-ReLU}\\
{\small Residual unit: Conv.-BN-ReLU-Dropout-Conv.,16 filters-\textcircled{+}-ReLU}\\
{\small Residual unit: Conv.-BN-ReLU-Dropout-Conv.,32 filters-\textcircled{+}-ReLU}\\
\end{tabular}
\begin{tabular}{p{8.6cm}}
{\small Residual unit: Conv.-BN-ReLU-Dropout-Conv.,32 filters-\textcircled{+}-ReLU}\\
{\small Residual unit: Conv.-BN-ReLU-Dropout-Conv.,32 filters-\textcircled{+}-ReLU}\\
{\small Residual unit: Conv.-BN-ReLU-Dropout-Conv.,64 filters-\textcircled{+}-ReLU}\\
{\small Residual unit: Conv.-BN-ReLU-Dropout-Conv.,64 filters-\textcircled{+}-ReLU}\\
{\small Residual unit: Conv.-BN-ReLU-Dropout-Conv.,64 filters-\textcircled{+}-ReLU}\\
{\small Global mean pooling}\\
{\small FC layer with \textbf{n} units where \textbf{n} is equal the number of action class.}\\
{\small Softmax layer}\\
\hline
\hline
\end{tabular}\\[3cm]
\begin{tabular}{p{8.6cm}}
\hspace*{2cm} {\small \textbf{Baseline 32-layer ResNet architecture}} \\
\hline
\hline
{\small 3x3 Conv., 16 filters, BN, ReLU}\\
{\small Residual unit: Conv.-BN-ReLU-Dropout-Conv.,16 filters-\textcircled{+}-ReLU}\\
{\small Residual unit: Conv.-BN-ReLU-Dropout-Conv.,16 filters-\textcircled{+}-ReLU}\\
{\small Residual unit: Conv.-BN-ReLU-Dropout-Conv.,16 filters-\textcircled{+}-ReLU}\\
{\small Residual unit: Conv.-BN-ReLU-Dropout-Conv.,16 filters-\textcircled{+}-ReLU}\\
{\small Residual unit: Conv.-BN-ReLU-Dropout-Conv.,16 filters-\textcircled{+}-ReLU}\\
{\small Residual unit: Conv.-BN-ReLU-Dropout-Conv.,32 filters-\textcircled{+}-ReLU}\\
{\small Residual unit: Conv.-BN-ReLU-Dropout-Conv.,32 filters-\textcircled{+}-ReLU}\\
{\small Residual unit: Conv.-BN-ReLU-Dropout-Conv.,32 filters-\textcircled{+}-ReLU}\\
{\small Residual unit: Conv.-BN-ReLU-Dropout-Conv.,32 filters-\textcircled{+}-ReLU}\\
{\small Residual unit: Conv.-BN-ReLU-Dropout-Conv.,32 filters-\textcircled{+}-ReLU}\\
{\small Residual unit: Conv.-BN-ReLU-Dropout-Conv.,64 filters-\textcircled{+}-ReLU}\\
{\small Residual unit: Conv.-BN-ReLU-Dropout-Conv.,64 filters-\textcircled{+}-ReLU}\\
{\small Residual unit: Conv.-BN-ReLU-Dropout-Conv.,64 filters-\textcircled{+}-ReLU}\\
{\small Residual unit: Conv.-BN-ReLU-Dropout-Conv.,64 filters-\textcircled{+}-ReLU}\\
{\small Residual unit: Conv.-BN-ReLU-Dropout-Conv.,64 filters-\textcircled{+}-ReLU}\\
{\small Global mean pooling}\\
{\small FC layer with \textbf{n} units where \textbf{n} is equal the number of action class.}\\
{\small Softmax layer}\\
\hline
\hline
\end{tabular}\\[0.75cm]
\\
\begin{tabular}{p{8.6cm}}
\hspace*{2cm} {\small \textbf{Baseline 44-layer ResNet architecture}} \\
\hline
\hline
{\small 3x3 Conv., 16 filters, BN, ReLU}\\
{\small Residual unit: Conv.-BN-ReLU-Dropout-Conv.,16 filters-\textcircled{+}-ReLU}\\
{\small Residual unit: Conv.-BN-ReLU-Dropout-Conv.,16 filters-\textcircled{+}-ReLU}\\
{\small Residual unit: Conv.-BN-ReLU-Dropout-Conv.,16 filters-\textcircled{+}-ReLU}\\
{\small Residual unit: Conv.-BN-ReLU-Dropout-Conv.,16 filters-\textcircled{+}-ReLU}\\
{\small Residual unit: Conv.-BN-ReLU-Dropout-Conv.,16 filters-\textcircled{+}-ReLU}\\
{\small Residual unit: Conv.-BN-ReLU-Dropout-Conv.,16 filters-\textcircled{+}-ReLU}\\
{\small Residual unit: Conv.-BN-ReLU-Dropout-Conv.,16 filters-\textcircled{+}-ReLU}\\
{\small Residual unit: Conv.-BN-ReLU-Dropout-Conv.,32 filters-\textcircled{+}-ReLU}\\
{\small Residual unit: Conv.-BN-ReLU-Dropout-Conv.,32 filters-\textcircled{+}-ReLU}\\
{\small Residual unit: Conv.-BN-ReLU-Dropout-Conv.,32 filters-\textcircled{+}-ReLU}\\
{\small Residual unit: Conv.-BN-ReLU-Dropout-Conv.,32 filters-\textcircled{+}-ReLU}\\
{\small Residual unit: Conv.-BN-ReLU-Dropout-Conv.,32 filters-\textcircled{+}-ReLU}\\
\end{tabular}
\begin{tabular}{p{8.6cm}}
{\small Residual unit: Conv.-BN-ReLU-Dropout-Conv.,32 filters-\textcircled{+}-ReLU}\\
{\small Residual unit: Conv.-BN-ReLU-Dropout-Conv.,32 filters-\textcircled{+}-ReLU}\\
{\small Residual unit: Conv.-BN-ReLU-Dropout-Conv.,64 filters-\textcircled{+}-ReLU}\\
{\small Residual unit: Conv.-BN-ReLU-Dropout-Conv.,64 filters-\textcircled{+}-ReLU}\\
{\small Residual unit: Conv.-BN-ReLU-Dropout-Conv.,64 filters-\textcircled{+}-ReLU}\\
{\small Residual unit: Conv.-BN-ReLU-Dropout-Conv.,64 filters-\textcircled{+}-ReLU}\\
{\small Residual unit: Conv.-BN-ReLU-Dropout-Conv.,64 filters-\textcircled{+}-ReLU}\\
{\small Residual unit: Conv.-BN-ReLU-Dropout-Conv.,64 filters-\textcircled{+}-ReLU}\\
{\small Residual unit: Conv.-BN-ReLU-Dropout-Conv.,64 filters-\textcircled{+}-ReLU}\\
{\small Global mean pooling}\\
{\small FC layer with \textbf{n} units, where \textbf{n} is equal the number of action class.}\\
{\small Softmax layer}\\
\hline
\hline
\end{tabular}\\[0.75cm]
\\
\begin{tabular}{p{8.6cm}}
\hspace*{2cm} {\small \textbf{Baseline 56-layer ResNet architecture}} \\
\hline
\hline
{\small 3x3 Conv., 16 filters, BN, ReLU}\\
{\small Residual unit: Conv.-BN-ReLU-Dropout-Conv.,16 filters-\textcircled{+}-ReLU}\\
{\small Residual unit: Conv.-BN-ReLU-Dropout-Conv.,16 filters-\textcircled{+}-ReLU}\\
{\small Residual unit: Conv.-BN-ReLU-Dropout-Conv.,16 filters-\textcircled{+}-ReLU}\\
{\small Residual unit: Conv.-BN-ReLU-Dropout-Conv.,16 filters-\textcircled{+}-ReLU}\\
{\small Residual unit: Conv.-BN-ReLU-Dropout-Conv.,16 filters-\textcircled{+}-ReLU}\\
{\small Residual unit: Conv.-BN-ReLU-Dropout-Conv.,16 filters-\textcircled{+}-ReLU}\\
{\small Residual unit: Conv.-BN-ReLU-Dropout-Conv.,16 filters-\textcircled{+}-ReLU}\\
{\small Residual unit: Conv.-BN-ReLU-Dropout-Conv.,16 filters-\textcircled{+}-ReLU}\\
{\small Residual unit: Conv.-BN-ReLU-Dropout-Conv.,16 filters-\textcircled{+}-ReLU}\\
{\small Residual unit: Conv.-BN-ReLU-Dropout-Conv.,32 filters-\textcircled{+}-ReLU}\\
{\small Residual unit: Conv.-BN-ReLU-Dropout-Conv.,32 filters-\textcircled{+}-ReLU}\\
{\small Residual unit: Conv.-BN-ReLU-Dropout-Conv.,32 filters-\textcircled{+}-ReLU}\\
{\small Residual unit: Conv.-BN-ReLU-Dropout-Conv.,32 filters-\textcircled{+}-ReLU}\\
{\small Residual unit: Conv.-BN-ReLU-Dropout-Conv.,32 filters-\textcircled{+}-ReLU}\\
{\small Residual unit: Conv.-BN-ReLU-Dropout-Conv.,32 filters-\textcircled{+}-ReLU}\\
{\small Residual unit: Conv.-BN-ReLU-Dropout-Conv.,32 filters-\textcircled{+}-ReLU}\\
{\small Residual unit: Conv.-BN-ReLU-Dropout-Conv.,32 filters-\textcircled{+}-ReLU}\\
{\small Residual unit: Conv.-BN-ReLU-Dropout-Conv.,32 filters-\textcircled{+}-ReLU}\\
{\small Residual unit: Conv.-BN-ReLU-Dropout-Conv.,64 filters-\textcircled{+}-ReLU}\\
{\small Residual unit: Conv.-BN-ReLU-Dropout-Conv.,64 filters-\textcircled{+}-ReLU}\\
{\small Residual unit: Conv.-BN-ReLU-Dropout-Conv.,64 filters-\textcircled{+}-ReLU}\\
{\small Residual unit: Conv.-BN-ReLU-Dropout-Conv.,64 filters-\textcircled{+}-ReLU}\\
{\small Residual unit: Conv.-BN-ReLU-Dropout-Conv.,64 filters-\textcircled{+}-ReLU}\\
\end{tabular}
\begin{tabular}{p{8.6cm}}
{\small Residual unit: Conv.-BN-ReLU-Dropout-Conv.,64 filters-\textcircled{+}-ReLU}\\
{\small Residual unit: Conv.-BN-ReLU-Dropout-Conv.,64 filters-\textcircled{+}-ReLU}\\
{\small Residual unit: Conv.-BN-ReLU-Dropout-Conv.,64 filters-\textcircled{+}-ReLU}\\
{\small Residual unit: Conv.-BN-ReLU-Dropout-Conv.,64 filters-\textcircled{+}-ReLU}\\
{\small Global mean pooling}\\
{\small FC layer with \textbf{n} units, where \textbf{n} is equal the number of action class.}\\
{\small Softmax layer}\\
\hline
\hline
\end{tabular}\\[0.75cm]
\\
\begin{tabular}{p{8.6cm}}
\hspace*{2cm} {\small \textbf{Baseline 110-layer ResNet architecture}} \\
\hline
\hline
{\small 3x3 Conv., 16 filters, BN, ReLU}\\
{\small Residual unit: Conv.-BN-ReLU-Dropout-Conv.,16 filters-\textcircled{+}-ReLU}\\
{\small Residual unit: Conv.-BN-ReLU-Dropout-Conv.,16 filters-\textcircled{+}-ReLU}\\
{\small Residual unit: Conv.-BN-ReLU-Dropout-Conv.,16 filters-\textcircled{+}-ReLU}\\
{\small Residual unit: Conv.-BN-ReLU-Dropout-Conv.,16 filters-\textcircled{+}-ReLU}\\
{\small Residual unit: Conv.-BN-ReLU-Dropout-Conv.,16 filters-\textcircled{+}-ReLU}\\
{\small Residual unit: Conv.-BN-ReLU-Dropout-Conv.,16 filters-\textcircled{+}-ReLU}\\
{\small Residual unit: Conv.-BN-ReLU-Dropout-Conv.,16 filters-\textcircled{+}-ReLU}\\
{\small Residual unit: Conv.-BN-ReLU-Dropout-Conv.,16 filters-\textcircled{+}-ReLU}\\
{\small Residual unit: Conv.-BN-ReLU-Dropout-Conv.,16 filters-\textcircled{+}-ReLU}\\
{\small Residual unit: Conv.-BN-ReLU-Dropout-Conv.,16 filters-\textcircled{+}-ReLU}\\
{\small Residual unit: Conv.-BN-ReLU-Dropout-Conv.,16 filters-\textcircled{+}-ReLU}\\
{\small Residual unit: Conv.-BN-ReLU-Dropout-Conv.,16 filters-\textcircled{+}-ReLU}\\
{\small Residual unit: Conv.-BN-ReLU-Dropout-Conv.,16 filters-\textcircled{+}-ReLU}\\
{\small Residual unit: Conv.-BN-ReLU-Dropout-Conv.,16 filters-\textcircled{+}-ReLU}\\
{\small Residual unit: Conv.-BN-ReLU-Dropout-Conv.,16 filters-\textcircled{+}-ReLU}\\
{\small Residual unit: Conv.-BN-ReLU-Dropout-Conv.,16 filters-\textcircled{+}-ReLU}\\
{\small Residual unit: Conv.-BN-ReLU-Dropout-Conv.,16 filters-\textcircled{+}-ReLU}\\
{\small Residual unit: Conv.-BN-ReLU-Dropout-Conv.,16 filters-\textcircled{+}-ReLU}\\
{\small Residual unit: Conv.-BN-ReLU-Dropout-Conv.,32 filters-\textcircled{+}-ReLU}\\
{\small Residual unit: Conv.-BN-ReLU-Dropout-Conv.,32 filters-\textcircled{+}-ReLU}\\
{\small Residual unit: Conv.-BN-ReLU-Dropout-Conv.,32 filters-\textcircled{+}-ReLU}\\
{\small Residual unit: Conv.-BN-ReLU-Dropout-Conv.,32 filters-\textcircled{+}-ReLU}\\
{\small Residual unit: Conv.-BN-ReLU-Dropout-Conv.,32 filters-\textcircled{+}-ReLU}\\
{\small Residual unit: Conv.-BN-ReLU-Dropout-Conv.,32 filters-\textcircled{+}-ReLU}\\
{\small Residual unit: Conv.-BN-ReLU-Dropout-Conv.,32 filters-\textcircled{+}-ReLU}\\
{\small Residual unit: Conv.-BN-ReLU-Dropout-Conv.,32 filters-\textcircled{+}-ReLU}\\
{\small Residual unit: Conv.-BN-ReLU-Dropout-Conv.,32 filters-\textcircled{+}-ReLU}\\
{\small Residual unit: Conv.-BN-ReLU-Dropout-Conv.,32 filters-\textcircled{+}-ReLU}\\
\end{tabular}
\begin{tabular}{p{8.6cm}}
{\small Residual unit: Conv.-BN-ReLU-Dropout-Conv.,32 filters-\textcircled{+}-ReLU}\\
{\small Residual unit: Conv.-BN-ReLU-Dropout-Conv.,32 filters-\textcircled{+}-ReLU}\\
{\small Residual unit: Conv.-BN-ReLU-Dropout-Conv.,32 filters-\textcircled{+}-ReLU}\\
{\small Residual unit: Conv.-BN-ReLU-Dropout-Conv.,32 filters-\textcircled{+}-ReLU}\\
{\small Residual unit: Conv.-BN-ReLU-Dropout-Conv.,32 filters-\textcircled{+}-ReLU}\\
{\small Residual unit: Conv.-BN-ReLU-Dropout-Conv.,32 filters-\textcircled{+}-ReLU}\\
{\small Residual unit: Conv.-BN-ReLU-Dropout-Conv.,32 filters-\textcircled{+}-ReLU}\\
{\small Residual unit: Conv.-BN-ReLU-Dropout-Conv.,32 filters-\textcircled{+}-ReLU}\\
{\small Residual unit: Conv.-BN-ReLU-Dropout-Conv.,64 filters-\textcircled{+}-ReLU}\\
{\small Residual unit: Conv.-BN-ReLU-Dropout-Conv.,64 filters-\textcircled{+}-ReLU}\\
{\small Residual unit: Conv.-BN-ReLU-Dropout-Conv.,64 filters-\textcircled{+}-ReLU}\\
{\small Residual unit: Conv.-BN-ReLU-Dropout-Conv.,64 filters-\textcircled{+}-ReLU}\\
{\small Residual unit: Conv.-BN-ReLU-Dropout-Conv.,64 filters-\textcircled{+}-ReLU}\\
{\small Residual unit: Conv.-BN-ReLU-Dropout-Conv.,64 filters-\textcircled{+}-ReLU}\\
{\small Residual unit: Conv.-BN-ReLU-Dropout-Conv.,64 filters-\textcircled{+}-ReLU}\\
{\small Residual unit: Conv.-BN-ReLU-Dropout-Conv.,64 filters-\textcircled{+}-ReLU}\\
{\small Residual unit: Conv.-BN-ReLU-Dropout-Conv.,64 filters-\textcircled{+}-ReLU}\\
{\small Residual unit: Conv.-BN-ReLU-Dropout-Conv.,64 filters-\textcircled{+}-ReLU}\\
{\small Residual unit: Conv.-BN-ReLU-Dropout-Conv.,64 filters-\textcircled{+}-ReLU}\\
{\small Residual unit: Conv.-BN-ReLU-Dropout-Conv.,64 filters-\textcircled{+}-ReLU}\\
{\small Residual unit: Conv.-BN-ReLU-Dropout-Conv.,64 filters-\textcircled{+}-ReLU}\\
{\small Residual unit: Conv.-BN-ReLU-Dropout-Conv.,64 filters-\textcircled{+}-ReLU}\\
{\small Residual unit: Conv.-BN-ReLU-Dropout-Conv.,64 filters-\textcircled{+}-ReLU}\\
{\small Residual unit: Conv.-BN-ReLU-Dropout-Conv.,64 filters-\textcircled{+}-ReLU}\\
{\small Residual unit: Conv.-BN-ReLU-Dropout-Conv.,64 filters-\textcircled{+}-ReLU}\\
{\small Residual unit: Conv.-BN-ReLU-Dropout-Conv.,64 filters-\textcircled{+}-ReLU}\\
{\small Global mean pooling}\\
{\small FC layer with \textbf{n} units, where \textbf{n} is equal the number of action class.}\\
{\small Softmax layer}\\
\hline
\hline
\end{tabular}
\end{document}